\documentclass[]{interact}

\usepackage{xcolor}
\usepackage{url}
\usepackage{multirow}
\usepackage{enumitem}
\usepackage{soul}
\usepackage{epstopdf}
\usepackage{subcaption}
\usepackage[sort]{natbib}
\bibpunct[, ]{(}{)}{;}{a}{,}{,}
\renewcommand\bibfont{\fontsize{10}{12}\selectfont}

\theoremstyle{plain}

\usepackage{array}

\theoremstyle{definition}

\theoremstyle{remark}

\begin{document}

\articletype{Research Article}

\title{Foundation Models for Geospatial Reasoning: Assessing the Capabilities of Large Language Models in Understanding Geometries and Topological Spatial Relations}

\author{\name{Yuhan Ji \textsuperscript{a}, Song Gao \textsuperscript{a,*}\thanks{*A preprint draft and the final version will be available on the International Journal of Geographical Information Science; Corresponding Author. Email: song.gao@wisc.edu}, Ying Nie \textsuperscript{a}, Ivan Majić \textsuperscript{b}, and Krzysztof Janowicz \textsuperscript{b,c}}
\affil{\textsuperscript{a} GeoDS Lab, Department of Geography, University of Wisconsin-Madison, USA;\\ \textsuperscript{b} Department of Geography and Regional Research, University of Vienna, Austria;\\ \textsuperscript{c} Department of Geography, University of California-Santa Barbara, USA}}

\maketitle

\begin{abstract}
AI foundation models have demonstrated some capabilities for the understanding of geospatial semantics. However, applying such pre-trained models directly to geospatial datasets remains challenging due to their limited ability to represent and reason with geographical entities, specifically vector-based geometries and natural language descriptions of complex spatial relations. To address these issues, we investigate the extent to which a well-known-text (WKT) representation of geometries and their spatial relations (e.g., topological predicates) are preserved during spatial reasoning when the geospatial vector data are passed to large language models (LLMs) including GPT-3.5-turbo, GPT-4, and DeepSeek-R1-14B. Our workflow employs three distinct approaches to complete the spatial reasoning tasks for comparison, i.e., geometry embedding-based, prompt engineering-based, and everyday language-based evaluation. Our experiment results demonstrate that both the embedding-based and prompt engineering-based approaches to geospatial question-answering tasks with GPT models can achieve an accuracy of over 0.6 on average for the identification of topological spatial relations between two geometries. Among the evaluated models, GPT-4 with few-shot prompting achieved the highest performance with over 0.66 accuracy on topological spatial relation inference. Additionally, GPT-based reasoner is capable of properly comprehending inverse topological spatial relations and including an LLM-generated geometry can enhance the effectiveness for geographic entity retrieval. GPT-4 also exhibits the ability to translate certain vernacular descriptions about places into formal topological relations, and adding the geometry-type or place-type context in prompts may improve inference accuracy, but it varies by instance. The performance of these spatial reasoning tasks unveils the strengths and limitations of the current LLMs in the processing and comprehension of geospatial vector data and offers valuable insights for the refinement of LLMs with geographical knowledge towards the development of geo-foundation models capable of geospatial reasoning.
\end{abstract}

\begin{keywords}
GeoAI; geospatial reasoning; large language models; GPT; topological spatial relations
\end{keywords}

\section{Introduction}
\label{sec:intro}
    
Our interaction with Artificial Intelligence (AI) based systems is changing radically due to progress in generative Foundation Models (FM) and the conversational, natural-language-driven style of interaction with many of these models. While most prior AI models were developed with a limited range of downstream tasks in mind, foundation models aim to be general-purpose building blocks supporting a broad range of applications. Essentially, they are trained on a substantially broader set of data and, while giving up accuracy for any specific task during development, are easily fine-tuned before or during deployment. Large language models (LLMs)~\citep{radford2019language, brown2020language}, such as Generative Pre-trained Transformers (GPT)~\citep{radford2018improving, achiam2023gpt}, and text-to-image models~\citep{frolov2021adversarial}, such as DALL-E~\citep{dalle}, are specific types of foundation models. Most of these models are generative, i.e., they return novel, synthetic output such as natural language answers or imagery instead of providing answers by (information) retrieval as was common in prior systems, e.g., from the field of expert systems. While foundation models may not inherently prescribe a specific interaction style, they can be trained or fine-tuned for various types of interactions by carefully crafting the training dataset for the intended purpose. For example, OpenAI's Codex is trained using paired code examples and comments, enabling natural language instructions to guide code generation effectively~\citep{chen2021evaluating}. Similarly, Contrastive language-image pre-training (CLIP) facilitates tasks like image search from paired textual descriptions. Reinforcement learning with human feedback (RLHF) is another approach that aligns model outputs with user intent, improving conversational dialog flow, adherence to prompts, and reducing harmful content. The resulting conversational style of interaction is part of their broad appeal but also causes new challenges. 

Together, these breakthroughs have opened the door towards conversation-style artificial GIS analysts (``GeoMachina'')~\citep{janowicz2020geoai}. For instance, ChatGPT-4 can understand instructions for frequent GIS tasks like reading in a dataset~\citep{mooney2023towards}, performing simple spatial analysis steps (by generating PySAL code), or even suggesting appropriate next steps. Consequently, researchers started exploring the capabilities and limits of current AI in representing spatial data~\citep{ji2023evaluating}, generating maps~\citep{zhang2023ethics}, extracting place semantics~\citep{hu2023geo}, automating GIS operations~\citep{li2023autonomous,zhang2024geogpt}, generating code~\citep{gramacki2024evaluation}, and drawing inferences from such data \citep{mai2023opportunities}. Interestingly, the gaps this early research revealed are not unexpected as they have been documented as pain points of prior AI systems before \citep{janowicz2015data}. Prominently featured among these shortcomings is the representation of and reasoning with topological spatial relations~\citep{cohn2008qualitative}. Even more, this is true across foundation models, i.e., LLMs and text-to-image models struggle similarly. For instance, ChatGPT~\citep{openai_chatgpt} will provide a metric distance (e.g., several kilometers) when asked about the border of two neighboring countries. Similarly, DALL-E frequently fails to generate images of regions or parts described using terms such as bordering, adjacent, contained, or specific types of maps~\citep{zhang2023ethics}. This is a critical insight as it implies that current work on geo-foundation models~\citep{xie2023geo}, e.g., location embeddings~\citep{mai2022review}, may benefit the broader AI community across models. 

To better understand the limitations of LLMs in handling spatial data and to develop foundation models for advancing geospatial artificial intelligence (GeoAI)~\citep{gao2023handbook}, this work aims to explore the potential of representing spatial object geometries in the WKT format to enable LLMs to perform GIS operations and enhance geospatial reasoning. In this work, we present intensive experiments with well-known text (WKT) representation of geometries as inputs for LLMs and with natural language descriptions of (vague) spatial configurations. However, it is important to note that, unlike other types of data, accurate geometries (e.g., points, polylines, and polygons) and their spatial relations, as used in GIS, are not usually expressed in natural language text for such models to consume during training. Without explicitly addressing such structural deficiencies, the proposed approach is not suggested to be directly applicable in practice.

The research contributions (RC) of our work are as follows: 
\begin{itemize}
\item RC1: We develop a workflow to assess the ability of LLMs to reason with topological spatial relations, more specifically, a subset of topological relations specified according to the Dimensionally Extended 9-Intersection Model (DE-9IM). To do so, we will compare two approaches. First, we will encode the geometries and their topological relations in an embedding space using LLMs. Second, we will use a prompt engineering method to pass WKT format of geometries directly to the LLMs.

\item RC2: To test the capabilities of LLMs, we firstly utilize the WKT representation of two geometries to predict the topological spatial relation between them, and then we use one of the geometries and the topological spatial relation to predict the second geometry. To do so, we will utilize the pre-trained text embedding models and also use prompt engineering to elicit the target geometry.

\item RC3: Finally, we study the ability of LLMs to extract the formalized topological spatial relations between geographic entities from vernacular descriptions (i.e., everyday language) of the relations between geographic entities, e.g., as found in administrative place descriptions from DBpedia/Wikipedia.

\end{itemize}

The remaining paper is organized as follows. We first review the literature on spatial relations, parts of the qualitative spatial reasoning and conceptual neighborhoods, large language models, and GeoAI foundation models in Section \ref{sec:review}. We then introduce the methodology and workflow used in this research in Section \ref{sec:method}, followed by the experiments design and dataset processing in Section \ref{sec:experiments}. After that, we present the experiment results about topological spatial relation qualification and retrieval tasks using LLMs in Section \ref{sec:results}. We further discuss the the confusion between the topological predicates with their corresponding conceptual neighborhoods in Section \ref{sec:discussion}. Finally, we conclude this paper and offer insights into future work in Section \ref{sec:conclusion}.

\section{Related work} \label{sec:review}
\subsection{Spatial relations}
Spatial relations refer to the connection between spatial objects regarding their geometric properties~\citep{guo1998spatial}, which specify the location of one object related to another one~\citep{carlson2001using} or more other objects~\citep{majic2021}. On the one hand, describing spatial relations in natural language is essential for understanding our surroundings in spatial cognition and navigating through space~\citep{freksa1998spatial}. On the other hand, a reverse parsing process, where exact spatial relations are identified from natural language descriptions, is vital to improving the quality of information retrieval and human-computer interaction in tasks such as map reading~\citep{head1984map}, geographic question answering~\citep{gao2013asking, mai2020se,scheider2021geo},  spatial query and reasoning~\citep{wang2000fuzzy, du2005spatial,guo2022deepssn}, disaster management~\citep{wang2016spatial, cervone2016using}, driving and robotics navigation~\citep{wallgrun2014building, tellex2011understanding}. 

Typically, binary spatial relations use the format of a triplet \textit{\{subject, predicate (preposition), object\}} to describe the relative positions of objects in space. In this format, the \textit{subject} is an entity being described in relation to another entity, the \textit{predicate (preposition)} is the descriptor between the subject and object, and the \textit{object} is the entity that the subject is being related to in terms of position or location. For example, ``Santa Barbara is situated northwest of Los Angeles'' would be expressed as \{Santa Barbara, northwest of, Los Angeles\} in the format of spatial relations. Even though spatial relations pervade in our daily life conversations, people tend to frequently use a limited number of predicates to describe \textit{topological}, \textit{directional}, and \textit{distance} relations~\citep{mark1994modeling,frank1992qualitative}. These expressions are qualitative in nature, offering approximate descriptions of an infinite range of possible spatial configurations. Nevertheless, speakers can convey complex spatial layouts by combining these basic predicates with contextual cues. For example, we might describe the locale of Santa Barbara as ``Santa Barbara is connected via U.S. Highway 101 to Los Angeles about 100 miles to the southeast.'', or the position of a person as standing ``in front of the building, facing east.'' The ability to combine and modify spatial predicates allows us to express a wide range of spatial relationships with a relatively small vocabulary but increases the difficulty of representing and understanding the meanings of such spatial relation descriptions for computers. The flexibility and ambiguity inherent in natural language often obscure the precise geometry of spatial arrangements, creating a disconnection between semantic interpretation and physical spatial layout.
The abundance of web documents containing geographical references offers the opportunity to retrieve spatially-aware information and support qualitative spatial reasoning from natural language texts~\citep{jones2004spirit}. To bridge the semantic-physical gap, prior work has focused on extracting spatial relations between named geographic entities by interpreting linguistic cues in text. These efforts include parsing grammatical and spatial semantic structures~\citep{kordjamshidi2011spatial, skoumas2016location, loglisci2012unsupervised}, as well as applying supervised machine learning models trained on annotated data with spatial linguistic features~\citep{yuan2011extracting, wu2023deep}. The resulting qualitative spatial relations, enriched by contextual narratives~\citep{wallgrun2015towards}, provide a foundation for computational models that link natural language semantics to the structured representations of physical space.

\subsection{Formalism of topological relations and conceptual neighborhoods}

In the field of GIS, attempts have been made to formalize the conversion between quantitative computational models of spatial relations and qualitative spatial representations from human discourse \citep{cohn2001qualitative, chen2015survey}. In \cite{clementini1994modelling}, topological relations are defined as spatial relations that are preserved under such transformations as rotation, scaling, and rubber sheeting. For topological spatial relations, region connection calculus (RCC)~\citep{randell1992spatial} and point-set topology intersection models (IM), e.g., 4-IM based on intersections of the \textit{boundaries} and \textit{interiors} of two objects~\citep{Egenhofer1991}, and 9-IM which also considers the \textit{exteriors} of two objects~\citep{Egenhofer1990}, are widely used approaches. RCC-8 ~\citep{cui1993qualitative} is a set of eight jointly exhaustive and pairwise disjoint relations defined for regions. The basic relations include topological predicates: equal (EQ), externally connected (EC), disconnected (DC), partially overlaps (PO), tangential (TPP/TPPi) and nontangential (NTPP/NTPPi) relations, which have been shown to be cognitively adequate to be well distinguished by humans~\citep{renz1998spatial}.

Point-set topology intersection models analyze whether intersections between the interiors, boundaries, and exteriors of two objects are empty or non-empty point sets.
The Dimensionally Extended 9-intersection model (DE-9IM) \citep{clementini1993small} further considered the dimensionality of each geometry in the intersection matrix so that the 9-IM is not a binary operation of intersects. Based on the DE-9IM model, five mutually exclusive relations are identified~\citep{clementini1996model}, including \{disjoint, touches (meets), crosses, overlaps, within\}. 
the Open Geospatial Consortium (OGC) later added \{intersects, contains, equals\} to the set for the convenience of GIS software users, and included in the \textit{GeoPandas} Python package for programmers. The recent development of RCC$^{*}$-9 expands the dimensions of RCC-8 and allows for a unified framework to model topological spatial relations ~\citep{clementini2014rcc, clementini2024extension}. However, since DE-9IM predicates were selected for better user interaction and have been implemented by OGC, this work focuses on DE-9IM. In \cite{mark1994modeling}, human subject testing was conducted to evaluate their model for spatial relations between lines and regions. The participants were presented with pairs of lines or regions and were asked to rate the spatial relation between them using a Likert scale that ranged from ``no relation" to ``strongly related". The pairs of lines and regions were generated based on the 19 topologically distinct spatial relations defined in the authors' model. The human judgments were then compared to the predicted spatial relations generated by their model. The results showed that the model's predicted topological spatial relations matched the human judgments with a high degree of accuracy, indicating the effectiveness of the model in capturing human perception of topological spatial relations.

\begin{figure}[h]
    \centering
    \begin{subfigure}[b]{0.45\textwidth}
        \centering
        \includegraphics[width=\textwidth]{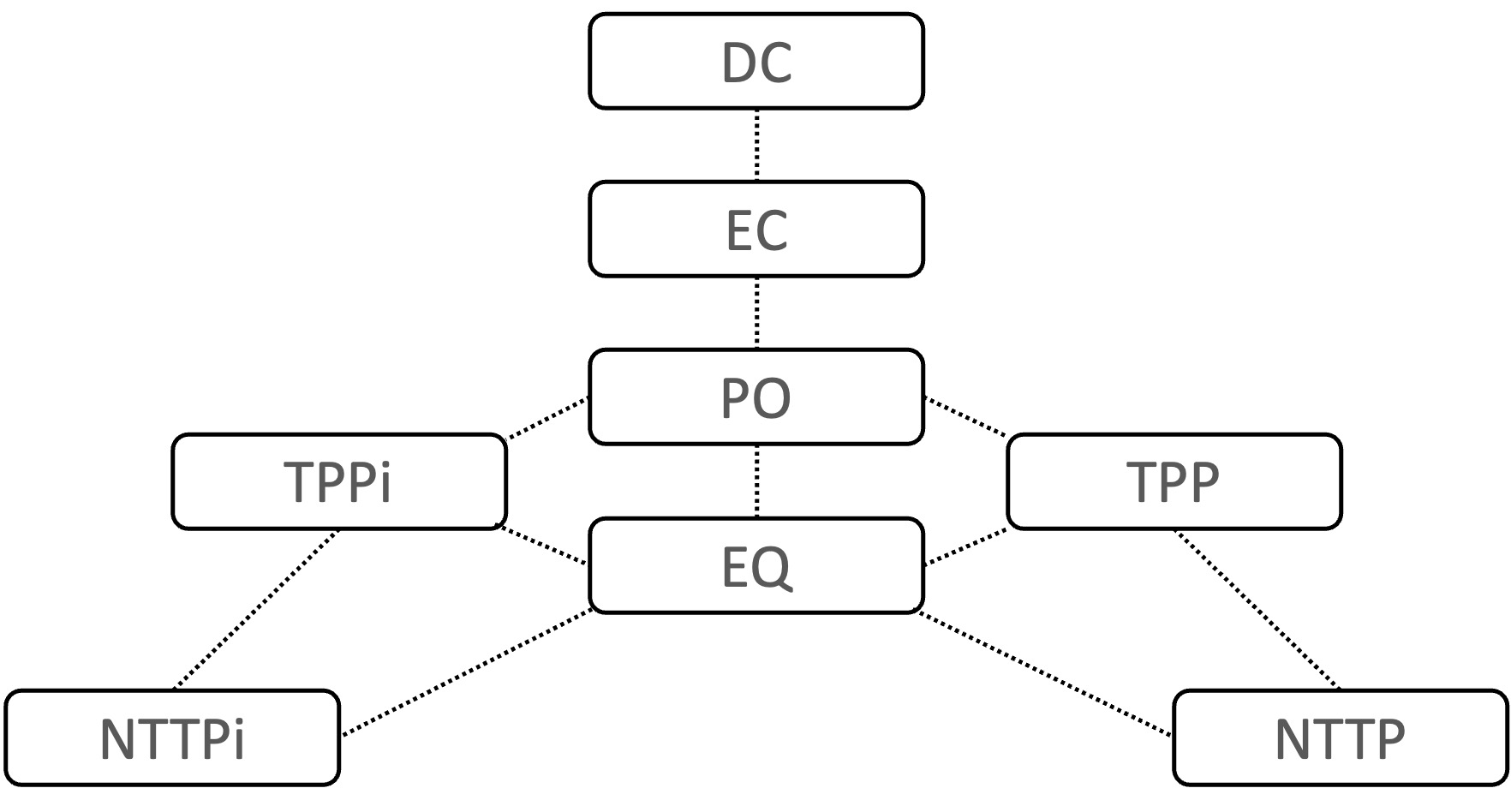} 
        \caption{}
        \label{fig:conceptual_neighborhood_a}
    \end{subfigure}
    \hspace{2em}
    \begin{subfigure}[b]{0.45\textwidth}
        \centering
        \includegraphics[width=\textwidth]{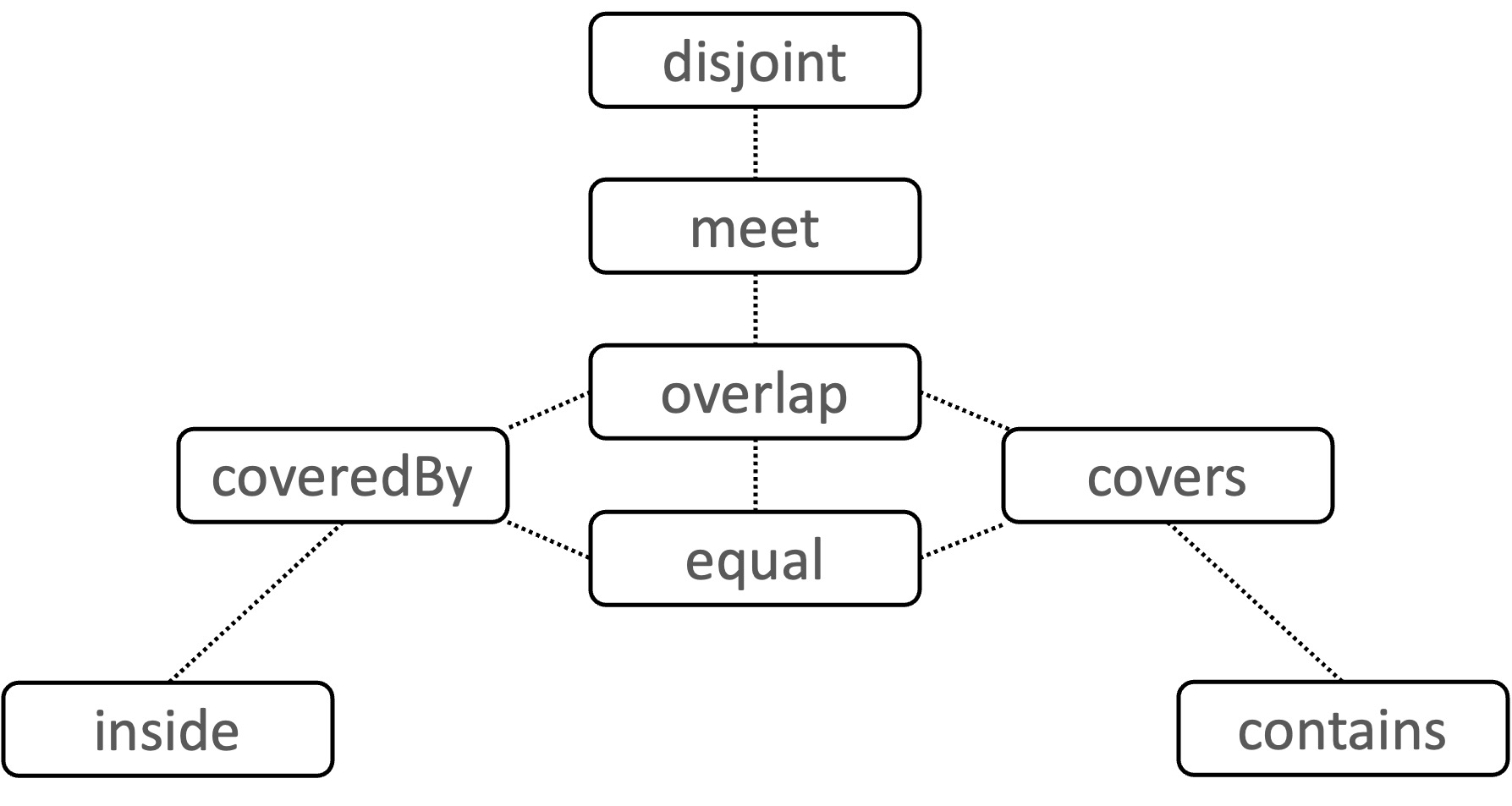} 
        \caption{}
        \label{fig:conceptual_neighborhood_b}
    \end{subfigure}
    \caption{The conceptual neighborhood of topological relations in RCC-8~\citep{randell1992spatial} on the left (redrawn for comparison) and 9-IM~\citep{egenhoferReasoningGradualChanges1992} on the right.}
\label{fig:conceptual_neighborhood}
\end{figure}

In both RCC and IM lineage, the idea of smooth transitions from one topological relation to another has been discussed early on. This means that, for example, if two polygon objects are disjoint, they would first require a touch relationship before moving to overlap. In this sense, some relationships are more similar or closer to each other than others, and this is known as the \textit{conceptual neighborhood} of topological relations. Figure~\ref{fig:conceptual_neighborhood} shows the neighborhood graphs using the RCC-8 (Figure \ref{fig:conceptual_neighborhood_a}) and 9-IM (Figure \ref{fig:conceptual_neighborhood_b}) nomenclature. Since the DE-9IM example only preserves the connection of a topological relation with its ``closest" relation, the inside/contains do not connect with equal in the graph. In addition to the conceptual neighborhood, \cite{egenhoferReasoningGradualChanges1992} proposed a formula for calculating the topological distance between topological relations using matrix representations, where the smaller distance means more similar between the two topological relations. We adopt the topological distance for evaluation later in this paper to provide a more nuanced perspective on whether LLMs' differentiation of topological relations aligns with human perception.

\subsection{Large language models and GeoAI foundation models}

The launch of ChatGPT by \cite{openai_chatgpt} marked a significant turning point, drawing widespread interest in Large Language Models (LLMs) and conversational AI from the public. Language-based foundation models boast an impressive range of parameters, from 110 million in BERT~\citep{devlin2018bert} to 1.5 billion in GPT-2~\citep{radford2019language}, and up to 137 billion in LaMDA (Google's Bard)~\citep{thoppilan2022lamda} and 175 billion in GPT-3~\citep{brown2020language}, demonstrating a significant variation in network architectures, scale, and purposes. Despite these differences, they share a common achievement: they have acquired a sophisticated understanding of language patterns and semantics, setting new performance standards in natural language processing tasks. Other types of foundation models include vision-based (e.g., vision transformer--ViT~\citep{dosovitskiy2020image} and segment anything model--SAM~\citep{kirillov2023segment}) and vision-language multimodal foundation models (e.g., Flamingo with 80 billion parameters~\citep{alayrac2022flamingo} and GPT-4 with over 1 trillion parameters~\citep{achiam2023gpt}). These pre-trained foundation models have been applied directly or transferred to a wide range of cross-domain tasks after fine-tuning or few-shot/zero-shot learning, e.g. education \citep{kasneci2023chatgpt}, healthcare ~\citep{yang2022large}, transportation ~\citep{zheng2023chatgpt}, etc. 

These foundation models have been trained on large-scale datasets that also contain geographical knowledge such as descriptions of locations and places in textual documents as well as spatial elements in maps, geo-referenced photos, and satellite imagery. Recently, researchers and institutions have begun the early exploration of integrating foundation models into GeoAI research and education. For example, \cite{mai2023opportunities} found that task-agnostic LLMs have the capability to surpass fully supervised deep learning models designed for specific tasks in understanding geospatial semantics, including toponym recognition, health data time-series forecasting, urban function, and scene classifications. 
\cite{hu2023geo} fused a few geo-knowledge examples into GPT models to improve the extraction of location descriptions from disaster-related social media messages. 
\cite{manvi2023geollm} found that geospatial knowledge can be effectively extracted from LLMs with auxiliary map data from OpenStreetMap. Additionally, spatial-context-aware prompts with pre-trained visual-language models can improve the accuracy of urban land use classification and urban function inference~\citep{huang2024zero,wu2023mixed},
In GIS, evaluations have been conducted to assess the qualitative spatial reasoning capabilities of LLMs in identifying and reasoning spatial relations using symbolic representations of spatial objects, such as RCC-8~\citep{cohn2023evaluation, cohn2024can} and cardinal directions~\citep{cohn2024evaluating}. While LLMs perceive the spatial structure through sequences of textual input~\citep{yamada2023evaluating} and leverage commonsense reasoning during their inference process~\citep{cohn2023dialectical}, they also demonstrate human-like misconceptions and distortions about space~\citep{fulman2024distortions}. Several studies~\citep{mai2022towards, tucker2024systematic, fernandez2023core} have proposed integrating vector data as a backbone for spatial reasoning. GPT-4 has shown the capability to generate coordinates for outlines of countries, rivers, lakes, and continents that approximate their actual geographic locations~\citep{das2023evaluating}. In \cite{ji2023evaluating}, LLM-generated embeddings can preserve geometry types and some coordinate information in the WKT representation of geometries. However, performing qualitative spatial reasoning and executing spatial tasks from implicit textual descriptions involving coordinates remains a significant challenge~\citep{majic2024spatial}. In addition, geospatial analysis workflows and operations can be automated when combing LLMs with spatial analysis tools~\citep{li2023autonomous,zhang2024geogpt}. ChatGPT can even achieve a promising grade when taking an introduction to GIS exam~\citep{mooney2023towards}.  In the field of Cartography,
\cite{tao2023mapping} explored the use of ChatGPT-4 for creating thematic maps and mental maps with appropriate prompts. However, \cite{zhang2023ethics} pointed out the ethical concerns on AI-generated maps' inaccuracies, misleading information, unanticipated features, and reproducibility.
In August 2023, NASA and IBM released their GeoAI Foundation Model--Prithvi, which was trained on NASA's Earth Observation remote sensing imagery (i.e., the harmonized Landsat and Sentinel-2 satellite dataset)~\citep{jakubik2023foundation} and has been found to have a good performance and transferability on flood
inundation mapping~\citep{li2023assessment}.
Alongside such remarkable achievements, there are concerns that need to be addressed together with the development and advancement of foundation models for GeoAI and geosciences (i.e., Geo-Foundation Models), such as geographical bias, diversity, spatial heterogeneity, limited human annotations, sustainability, privacy and security risks~\citep{janowicz2023philosophical,xie2023geo,rao2023building,hu2024a}.

\section{Methodology}\label{sec:method}
\subsection{Preliminaries and Workflow}

\begin{figure}[h]
    \centering
    \includegraphics[width=.99\linewidth]{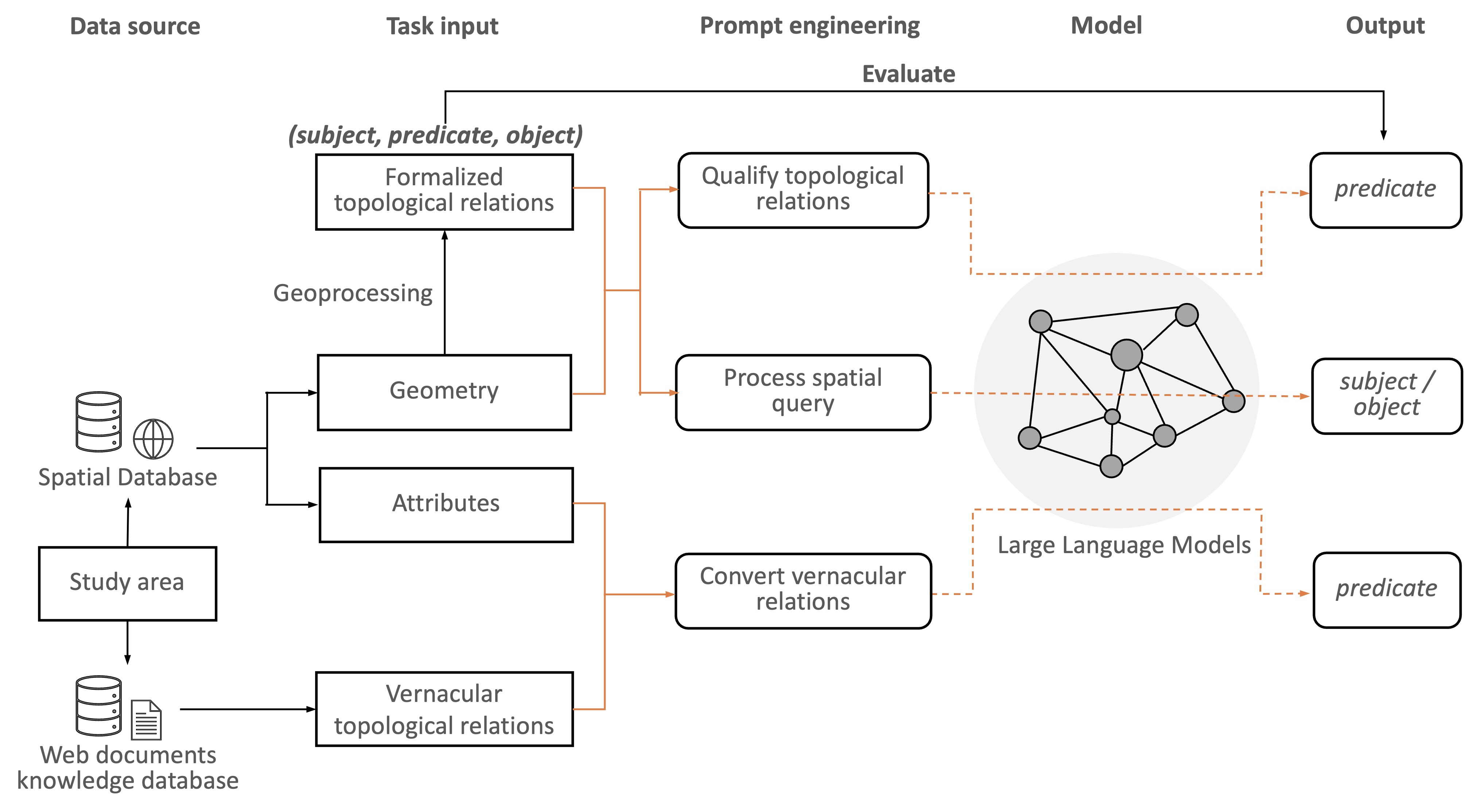}
    \caption{Overview of the workflow in this research.}
    \label{fig:framework}
\end{figure}

This research focuses on assessing the ability of LLMs to represent textual descriptions of geometries and understand topological spatial relations between geometric objects. The overall framework of this research is shown in Figure \ref{fig:framework}.
Given a study area, we first retrieve spatial objects from both a spatial database and a textual description about places from a Web document knowledge database (e.g., DBpedia/Wikipedia). When the documents contain vernacular 
 description of topological relations between two places, formalized DE-9IM topological spatial relations will be extracted from the spatial footprints (geometries) in the format of triplets as ground truth. The obtained geometric, attributive, and relational information is used as input for downstream tasks (e.g., qualify topological relations, process spatial query, and convert vernacular description of relations), where task-specific prompts are designed accordingly. The task output from the LLMs is then compared to the ground truth topological relation triplets to evaluate their ability to encode and reason about geometries and topological spatial relations.
The following subsections will further provide details on each evaluation task and the corresponding workflow. The definitions and notations used in this paper are listed in Table \ref{tab:notation}.

%


\begin{table}[!ht]
\footnotesize
    \centering
    \caption{Notations}
    \begin{tabular}{p{.2\linewidth}|p{.75\linewidth}}
    \hline
        \textbf{Notation} &  \textbf{Description}\\
        \hline
        $A/B$ & The objectID of spatial objects A or B \\
        $g_A$ & The geometry of $A$  that can be processed in GIS tools \\
        $GeomType(A)$ & The geometry type of $A$, (e.g. Point, LineString, and Polygon when $g_A$ is a simple feature) \\
        $g_A^{\circ}$ & The interior of $g_A$ \\
        $dim(g)$ & \begin{tabular}[m]{@{}l@{}} The dimension of a geometry $g$. \\ $dim(g)=\begin{cases}
            - & g=\emptyset \\
            0 & g \text{ contains at least one Point without Linestrings or Polygons} \\
            1 & g \text{ contains at least one Linestrings without Polygons} \\
            2 & g \text{ contains at least one Polygon}
        \end{cases}$ \end{tabular}\\
        $WKT(A)$ & The WKT format of $g_A$ \\
        $Enc(A)$ & The location encoding of $g_A$ using an LLM model to encode $WKT(A)$\\
        $R$ & The set of predicates to represent the topological spatial relations in this research,  i.e. \{equals, disjoint, crosses, touches, contains, within, overlaps\}, as defined by OGC and implemented in GeoPandas.\\
        $rel$ & A predicate that can be used to represent the topological spatial relation, $rel \in R$\\ 
        $Rel(A, B)$ & The topological spatial relation between the subject $A$ and the object $B$ \\
        $\left[Enc(A);Enc(B)\right]$ & The concatenation of the embeddings of $A$ and $B$ \\
        $D(rel_1, rel_2)$ & The topological distance between two relations $rel_1$ and $rel_2$ on the conceptual neighborhood graph \citep{egenhoferReasoningGradualChanges1992}\\
        $S_A(rel, B)$ & The relevancy score of a retrieved subject $A$ given the reference object $B$ and the desired topological spatial relation $rel$.\\
    \hline
    \end{tabular}
    \label{tab:notation}
\end{table}

\subsection{Determining topological spatial relations} \label{sec:spatialrelation}

\begin{figure}[h]
    \centering
\includegraphics[width=.85\linewidth]{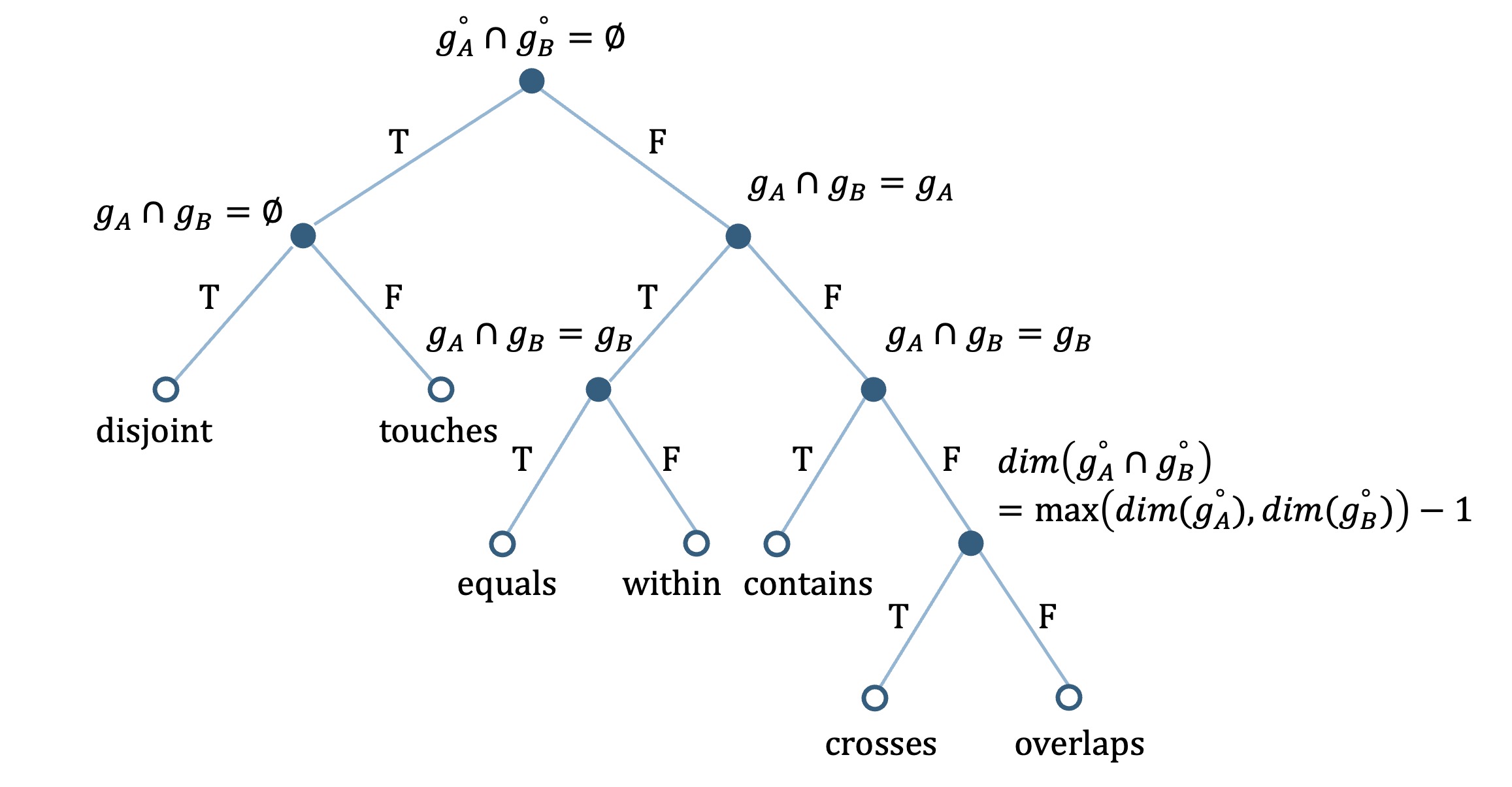}
    \caption{The decision tree for the topological spatial relations.}
    \label{fig:tree}
\end{figure}

In the original work of DE-9IM \citep{clementini1993small}, the five defined topological predicates \{disjoint, touches (meets), crosses, overlaps, within\} were considered mutually exclusive. However, the statement no longer holds with the introduction of ``contains" and ``equals" to the set by the OGC standard. 
Therefore, to ensure the uniqueness of the topological spatial relations between two objects, we interpret ``within" as ``within (but not equals)" and ``contains" as ``contains (not equals)" in this work. 
Accordingly, we modify the decision tree in \cite{clementini1993small} to do the reasoning about the topological relations between two spatial objects, as illustrated in Figure \ref{fig:tree}. Based on the decision process, the topological spatial relations do not apply to every combination of geometry types. The definitions and possible geometry type combinations of the seven predicates used in this research are listed in table \ref{tab:geom_comb}. Several visual examples of the topological spatial relations between two geometries can be seen in Figure \ref{fig:t2_relations}.

\begin{table}[h]
\footnotesize
    \centering
    \caption{The named topological spatial predicates with the 9-intersection Boolean code (T: true; F: false; *: free value) and corresponding applicable geometry type combinations of a predicate. }
    \begin{tabular}{>{\centering\arraybackslash}p{.3\linewidth}|p{.6\linewidth}}\hline
    \multicolumn{1}{c|} {\textbf{Predicate with 9-intersection code}} & \multicolumn{1}{c}{\textbf{Geometry Type Combination}}\\ \hline 
    equals: T*F**FFF*   & Point/Point, LineString/LineString, Polygon/Polygon \\\hline
    within: T*F**F***   & \begin{tabular}[m]{@{}l@{}}Point/LineString, Point/Polygon, \\LineString/LineString, LineString/Polygon, \\ Polygon/Polygon\end{tabular} \\\hline
    contains: T*****FF*  & \begin{tabular}[m]{@{}l@{}}LineString/Point, LineString/LineString, \\ Polygon/Point, Polygon/LineString, Polygon/Polygon\end{tabular}    \\\hline
    overlaps: T*T***T**  & LineString/LineString, Polygon/Polygon\\\hline
    touches: FT******* or F***T**** & \begin{tabular}[m]{@{}l@{}}Point/LineString, Point/Polygon, \\ LineString/Point, LineString/LineString, LineString/Polygon, \\ Polygon/Point, Polygon/LineString, Polygon/Polygon\end{tabular} \\\hline
    crosses: T*T******   & LineString/LineString, LineString/Polygon, Polygon/LineString \\\hline
    disjoint: FF*FF****	 & Applicable to ALL  \\\hline                  \end{tabular}

    \label{tab:geom_comb}
\end{table}

\subsection{Representing geospatial data as text}

An embedding is a multi-dimensional numeric vector representation of objects to capture the complex patterns and relationships in the data. While researchers have explored different approaches to embed geometries using spatially explicit models \citep{mai2022review, zhu2022reasoning, yan2017itdl}, this study presents a novel perspective by hypothesizing that LLMs can effectively encode the WKT format of geospatial vector data (points, polylines, and polygons) and preserve crucial geometric information. We adopt sentence embedding models~\citep{logeswaran2018efficient, reimers2019sentence, neelakantan2022text} to generate neural embeddings of the input geometry WKT strings, which allows for the comparison and retrieval of spatial information through the semantic search~\citep{muennighoff2022sgpt,hu2015metadata}.


%
%

\subsection{Evaluation Tasks}\label{sec:tasks}

\subsubsection{Topological spatial relation qualification}\label{sec:task1}


In \cite{wolter2012qualitative}, spatial relation qualification is defined as the process of inferring qualitative spatial relations from quantitative data. The first task aims to leverage LLMs to \textit{classify} the topological spatial relationships between subject entity A and object entity B into one of seven predefined topological predicates (in Section \ref{sec:spatialrelation}), combined with their geometry types. The input and output of \textbf{Task 1} are described as follows:

\textbf{Input}: The input for this task is the WKT representations of geometries A and B, denoted as WKT(A) and WKT(B). Example inputs:
\begin{itemize}[]
    \item $WKT(A)$: POINT (-89.3551 43.123) 
    \item $WKT(B)$: POLYGON ((-89.3552 43.124, -89.355 43.124, -89.355 43.122, -89.3552 43.122, -89.3552 43.124))
\end{itemize}

\textbf{Output}: The output is a tuple that describes the topological spatial relationship between the two geometries, in the format of $(GeomType(A), predicate, GeomType(B))$. Given the example inputs, the expected output of a correct classification would be:

\begin{itemize}[]
    \item (Point, within, Polygon)
\end{itemize}

\textbf{Use Case}:  Task 1 can be relevant to linking the geometries that occur in the same spatial context. For example, suppose one document already provides location, geometry and attribute information on housing resources and public transportation facilities. In that case, the LLM may directly use geographic information and other contexts to suggest affordable and accessible housing by public transportation facilities.


\begin{figure}[h]
    \centering
\includegraphics[width=.99\linewidth]{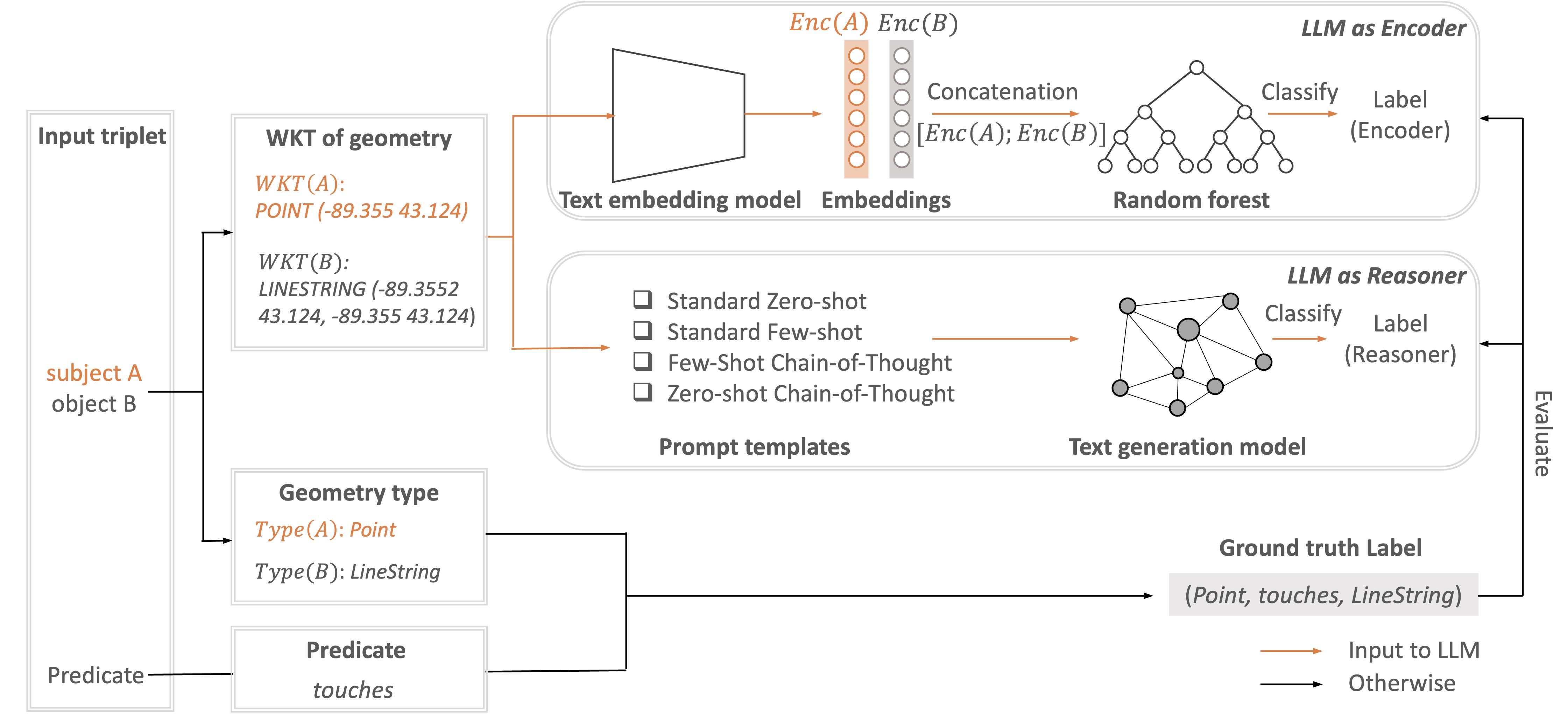}
    \caption{The workflow for the topological spatial relation qualification task.}
    \label{fig:task1}
\end{figure}

\begin{figure}[h]
    \centering
\includegraphics[width=1.1\linewidth]{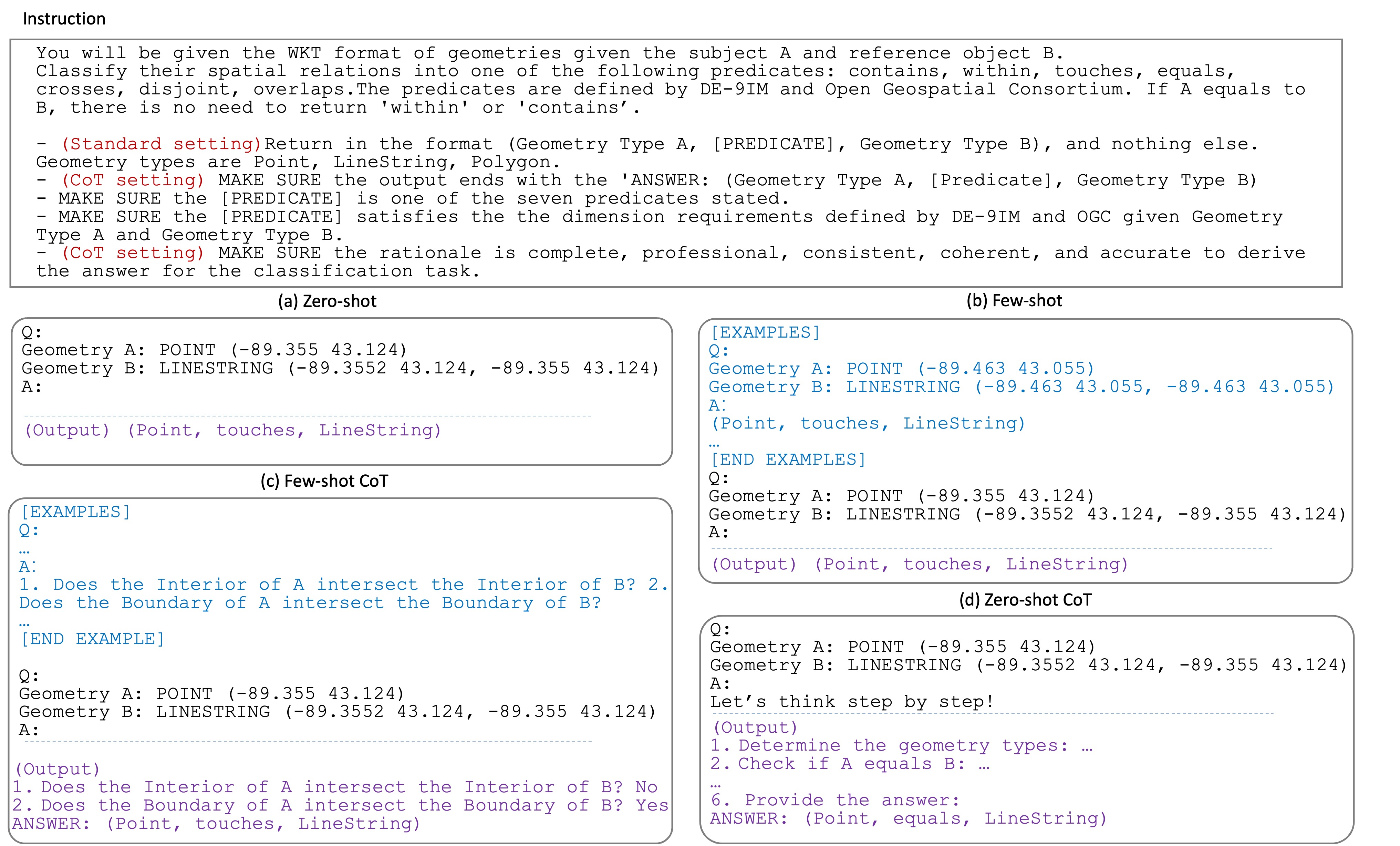}
    \caption{Topological spatial relation qualification example inputs and outputs with different prompt techniques.}
    \label{fig:prompt}
\end{figure}

The workflow of task 1 is shown in Figure \ref{fig:task1}. Given an input triplet that describes the topological spatial relation between subject A and object B, i.e., (subject, predicate, object), we first retrieve the WKT strings, and Geometry types of A and B. We then adopt two approaches (embedding-based and prompt-based) to perform the task, utilizing an appropriate LLM, to function as either a text encoder or a reasoner. For encoding, a pre-trained sentence embedding model generates the embeddings of the geometries of A and B. The embeddings are concatenated as the input for a random forest classifier~\citep{breiman2001random}. For reasoning, a more powerful generative model, such as GPT-4 and DeepSeek-R1, are employed to perform the task defined in the prompt. Four prompt engineering techniques are adopted to potentially guide the LLMs towards producing a more valid and accurate output of the topological spatial relation, including standard zero-shot learning, standard few-shot learning \citep{radford2019language}, few-shot chain-of-thought (CoT) prompting \citep{wei2022chain}, and zero-shot COT prompting\citep{kojima2022large}. In few-shot CoT, we follow the decision tree in Figure \ref{fig:tree} to generate the intermediate steps to determine the topological spatial relations as examples. While the identification of topological spatial relations might appear straightforward to the human brain, it involves multi-step reasoning. The DE-9IM framework~\citep{clementini1993small} decomposes the problem into intersections of the boundaries, interiors, and exteriors of two geographic entities, with dimensional requirements that map to topological predicates intuitive to users. We hypothesized that few-shot prompting and explicit reasoning steps, guided by CoT, could improve the model's performance on this qualification task. The example inputs and outputs of the topological spatial relation qualification task using the above-mentioned different prompt engineering techniques are illustrated in Figure \ref{fig:prompt}.

The metrics for evaluating the topological spatial relation qualification task are as follows. 

\begin{enumerate}[series=outerlist] 
    \item Validity
\end{enumerate}

    
    \begin{enumerate}[label=\alph*., series=innerlist]
    \item Valid format of the output: LLMs should follow the instructions to use the given format of the output in (Geometry Type A, Predicate, Geometry Type B).

    \item Valid geometry types: LLMs should preserve the Geometry Type A and Geometry Type B from the given WKT format of geometries.

    \item Valid combinations of geometry types for the topological predicates as shown in table \ref{tab:geom_comb}.
    
    \end{enumerate}
    
\begin{enumerate}[resume=outerlist] 
    \item Accuracy
\end{enumerate}
    
    For valid outputs, we can compute the accuracy when the output topological spatial predicate matches the ground truth. 
        
\begin{enumerate}[resume] 
    \item Topological distance in the conceptual neighborhood graph
\end{enumerate}
    
In this work, we use the shortest path distance between two topological predicates in the conceptual neighborhood graph (see Figure \ref{fig:conceptual_neighborhood}), where the distance of each edge equals 1. Since Figure \ref{fig:conceptual_neighborhood} was originally proposed for region-to-region (Polygon/Polygon) relations in 9IM, we mapped their topological predicates to the seven DE-9IM predicates that we use. For other geometry type combinations, we refer to \cite{mark1994modeling} and \cite{reis2008conceptual} to extract the conceptual neighborhood graphs. With the topological distance measurement, we can further analyze which pairs of predicates can easily confuse LLMs and whether such confusion is directed, by comparing the false-negative and false-positive results.

\subsubsection{Spatial query processing}


In \cite{sack1999handbook}, a generic spatial query is defined as the retrieval of subjects from a set of candidate geometric entities that are in a specific relation $rel$ with the query object $B$ on the basis of geometric information only. Our second task aims to evaluate whether LLMs can jointly encode a topological relation and one geometry to capture the feasible geometries that meet the query requirement. The input and output of the \textbf{Task 2} are as follows:

\textbf{Input}: The input for this task is the WKT representations of geometry $B$, denoted as $WKT(B)$, and a given predicate of topological spatial relations $rel$. Example input:

\begin{itemize}[]
    \item Predicate: within 
    \item $WKT(B)$: POLYGON ((-89.3552 43.124, -89.355 43.124, -89.355 43.122, -89.3552 43.122, -89.3552 43.124))
\end{itemize}

\textbf{Output}: The output is the identifier of a subject entity $A$ whose topological spatial relationship with B is described by the predicate.

\textbf{Use Case}: Task 2 is valuable for retrieving textual reports that involve locations, spatial layouts, and geospatial semantics. This analysis relies on accurate queries using spatial predicates. For instance, it would be beneficial to analyze the selection of a nearby competitor's site report when considering opening a business in the same neighborhood.

\begin{figure}[h]
    \centering
\includegraphics[width=.99\linewidth]{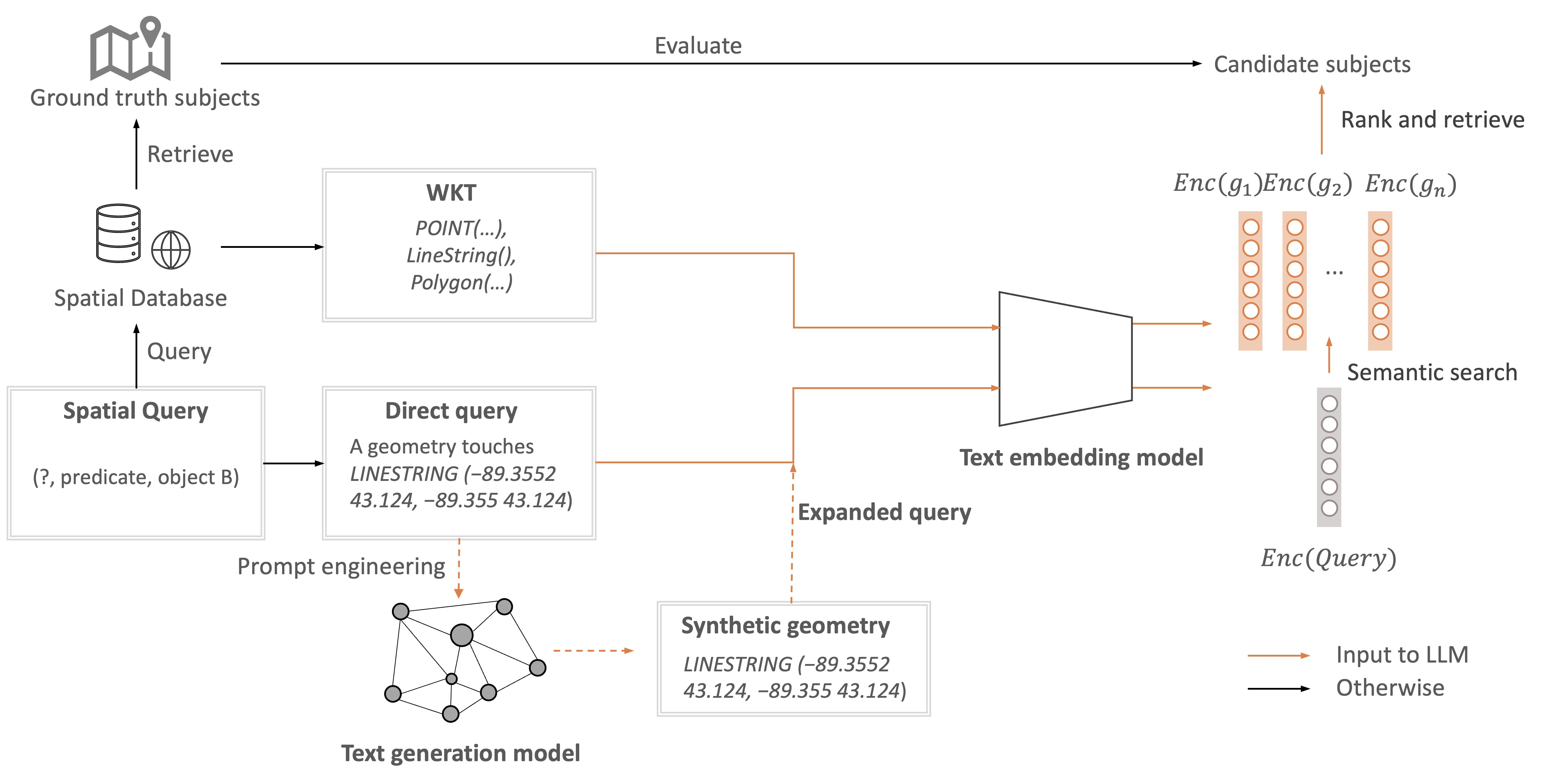}
    \caption{The workflow for the spatial query processing task.}
    \label{fig:task2}
\end{figure}

The evaluation workflow of Task 2 is shown in Figure \ref{fig:task2}.  Given a query specifying the topological spatial relation $rel$ with the query object $B$ ($WKT(B)$), we first retrieve the subjects from the study area spatial database as ground truth. We format the query as the input to an LLM using two approaches. First, this query can be directly formulated as a sentence, such as ``Retrieve a geometry  within POLYGON ((-89.3552 43.124, -89.355 43.124, -89.355 43.122, -89.3552 43.122, -89.3552 43.124)." Alternatively, synthetic geometries can be created using a generative model to expand the query, connecting the query with the search space. The (expanded) query text is inputted into the sentence embedding model to generate the embeddings. The geometries in WKT format for spatial entities are also processed by the same embedding model, to generate the embeddings $(Enc(g_1), Enc(g_2), ..., Enc(g_n))$. 
The most relevant subject geometries are retrieved based on the cosine similarity of their geometry embeddings and the query embeddings.  
We perform the evaluation as a link prediction task in the ``filtered'' setting \citep{bordes2013translating}, which excludes other subjects related to B by the topological predicate $rel$ from the database and concentrates on the retrieval of the subject in the triplet. This approach addresses the biases introduced by the significant difference in the number of spatially related subjects across predicates and the objects. Finally, the retrieved subjects are evaluated by their actual topological spatial relation to the reference subject.

In the following, we introduce how to format the direct query and the expanded query with LLM-generated geometries in detail:

\begin{enumerate}[] %
    \item Direct Query
\end{enumerate}
    
    Given the WKT format of the geometry of a known reference object (e.g., LINESTRING (-89.4534 43.035, -89.454 43.0351)) and a designated topological spatial relation (e.g., ``crosses"), the query formulation is as follows: ``Retrieve a geometry that crosses the LINESTRING (-89.4534 43.035, -89.454 43.0351)." If the search focuses on a specific geometry type, the query can be articulated as ``Retrieve a LINESTRING geometry that crosses the LINESTRING (-89.4534 43.035, -89.454 43.0351)." 

\begin{enumerate}[resume]    
    \item Expanded query with LLM-generated geometries
\end{enumerate}
    
\begin{figure}[h]
    \centering
\includegraphics[width=1.0\linewidth]{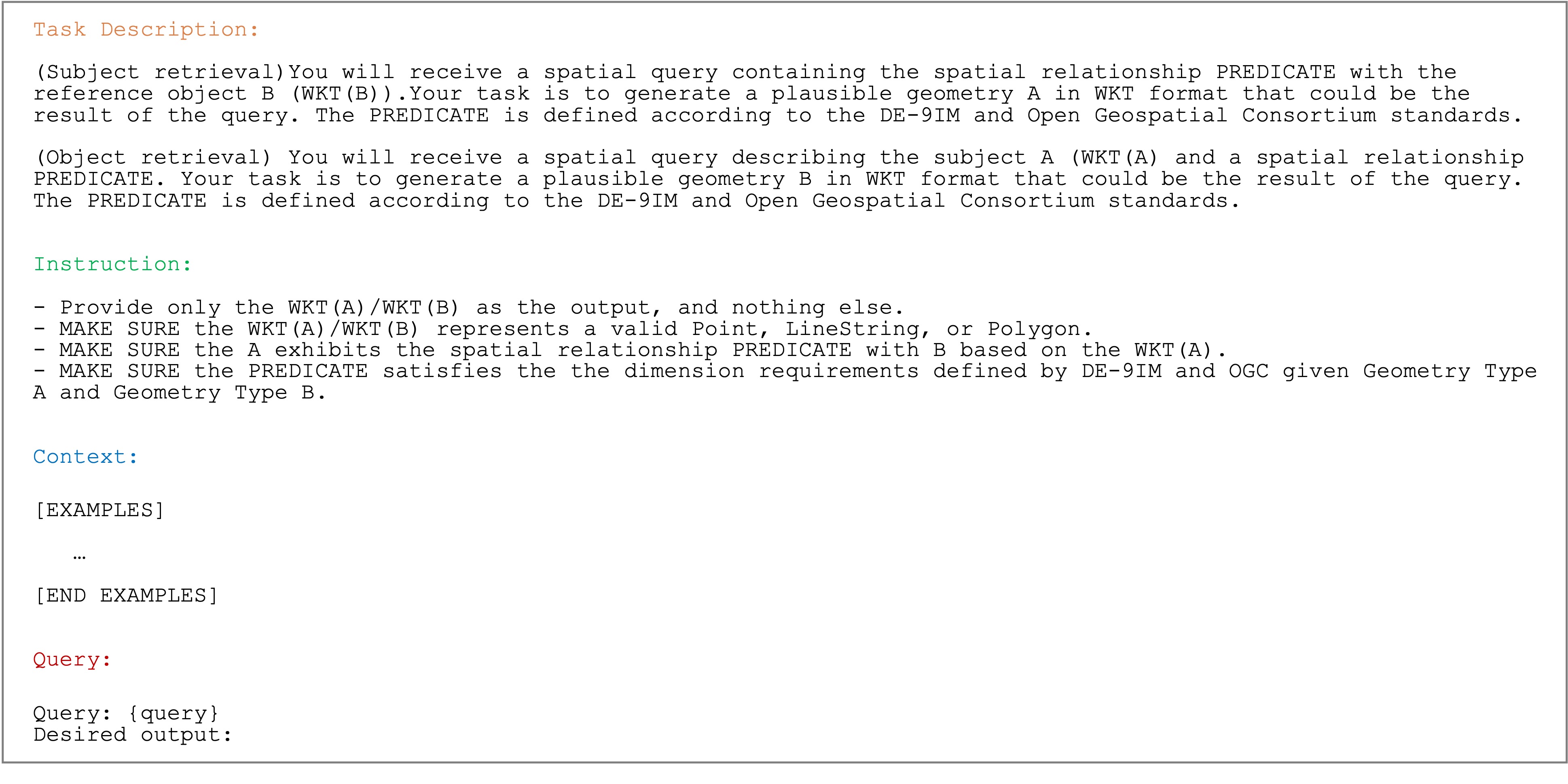}
    \caption{Prompt template used for geometry generation in the spatial query processing task.}
    \label{fig:t2_prompt}
\end{figure}

    In \cite{carpineto2012survey} and \cite{hu2015metadata}, (geospatial) query expansion is used to augment the user's original query with new features (e.g., geographic or thematic characteristics) that share a similar meaning as the expected output of semantic search. The method can address the lack of semantic similarity between the query and the desired geometry. We extend the \textit{Query2Doc} model~\citep{wang2023query2doc} to the spatial query expansion, where we leverage an LLM to generate a synthetic geometry that can possibly be the response to the query. The prompt template for the generation of geometric objects or subjects is listed in Figure \ref{fig:t2_prompt}. We adopt the following prompting approaches for geometry generation. 
    
    \begin{itemize}
        \item Zero-shot: LLMs generate geometries directly from the given spatial query.
        
        \item Zero-shot + Self-check: LLMs are asked to verify the spatial relations before generating the output.
    
        \item Few-shot: Give a few pairs of example queries and corresponding subjects while maintaining spatial relations and object geometry type.
    
        \item Few-shot + Negative examples: Apart from the plausible examples, we also include the negative examples that are not the correct responses for the given query. The examples are formatted as ``Retrieve a {Geometry Type} which ... Good Response:... Bad Response: ..."
    \end{itemize}

    We further incorporate the LLM-generated geometries into the spatial queries to assess the usefulness of the expanded queries.

The evaluation includes two parts: First, LLMs' ability to generate valid synthetic geometries as a basis for the expanded queries. Second, query processing performance through semantic search using both direct queries and expanded queries.

\begin{enumerate}[series=outerlist] 
    \item Validity of the LLM-generated geometries
\end{enumerate}


    \begin{enumerate}[label=\alph*., series=innerlist]
    \item Valid WKT format of geometries to be successfully parsed by the GIS tool for creating geometry instances.
    
    \item Correct topological spatial relation $Rel$ with the query object $B$. 
    \end{enumerate}
    
\begin{enumerate}[resume=outerlist] 
    \item Mean Reciprocal Rank (MRR) and Hits@K of the retrieval performance 
\end{enumerate}
    
    We employ two commonly used metrics in geographic information retrieval, Mean Reciprocal Rank (MRR)~\citep{yan2017itdl} and Hits@K, in the ``filtered" setting~\citep{bordes2013translating}. 
    A desirable model is characterized by higher MRR and Hits@K values.

\subsubsection{Conversion of vernacular relation descriptions}\label{sec:vernacularrelations}

In \cite{chen2018graph}, a vernacular description of spatial relations between places is an alternative to formal spatial relations in metric space, which occurs in everyday communication in a flexible format of a preposition, verb, phrase, or even implicit text description. The third task aims to evaluate how LLMs can convert the vernacular description (i.e., everyday language) of a topological relationship between two geographic entities into one of seven predefined topological predicates based on the given context. This task is inspired by LLM's commonsense model of the world and naive geographical knowledge about space, and the domain-specific knowledge of the formalism in calculus to bridge the gap between the vernacular(narrative) descriptions and the formalized topological predicates. For example, ChatGPT is able to provide the rationale behind the statement ``When an island is in the middle of a lake, the island touches the lake if the lake is considered as a separate region (not fully containing the island)" by identifying the lake in this scenario is a double-border object using commonsense knowledge reasoning (rather than precise geometries). It then maps this understanding to the “touches” topological relation, applying expertise in the GIS domain. The input and output of the \textbf{Task 3} are as follows:

\textbf{Input}: The input for this task includes a sentence that describes the topological relationship between two places in everyday language, along with the contextual information of the two places. Example input: 

\begin{itemize}[] 
    \item Sentence: Place $A$ is home to Place $B$   
    \item Context: Place $A$ is a city. Place $B$ is a university
\end{itemize}

\textbf{Output}: The output will rephrase the sentence using the formalized topological predicates.

\begin{itemize}[]
    \item Answer: Place $A$ contains Place $B$
\end{itemize}

\textbf{Use Case}: Parsing vernacular descriptions of spatial relations between places into formal ones can better support the users interacting in natural language and the use of spatial analysis tools that rely on formal topological relations. For example, interpreting vague terms in travel reports to determine if cross-border human behavior exists and interpreting the territorial changes and alignment of contemporary boundaries in the historical context.

\begin{figure}[h]
    \centering
\includegraphics[width=.99\linewidth]{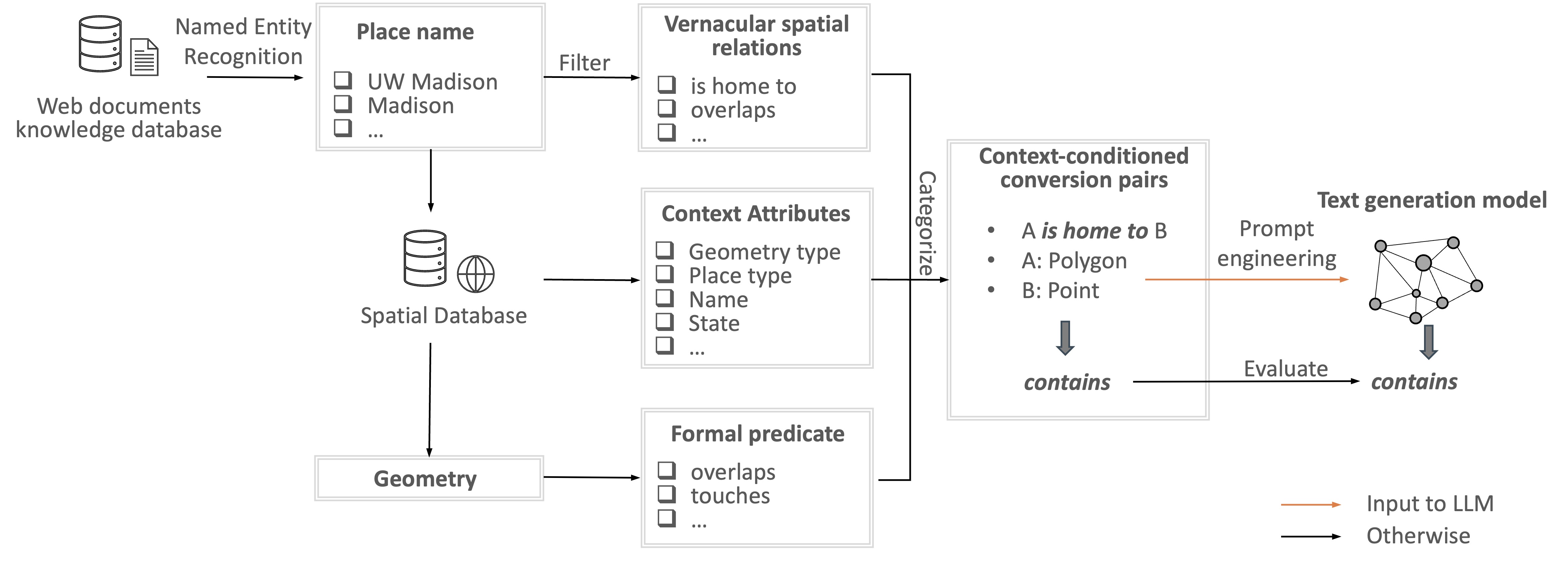}
    \caption{The workflow for the vernacular relations conversion task.}
    \label{fig:workflow_3}
\end{figure}

\begin{figure}[h]
    \centering
\includegraphics[width=1.0\linewidth]{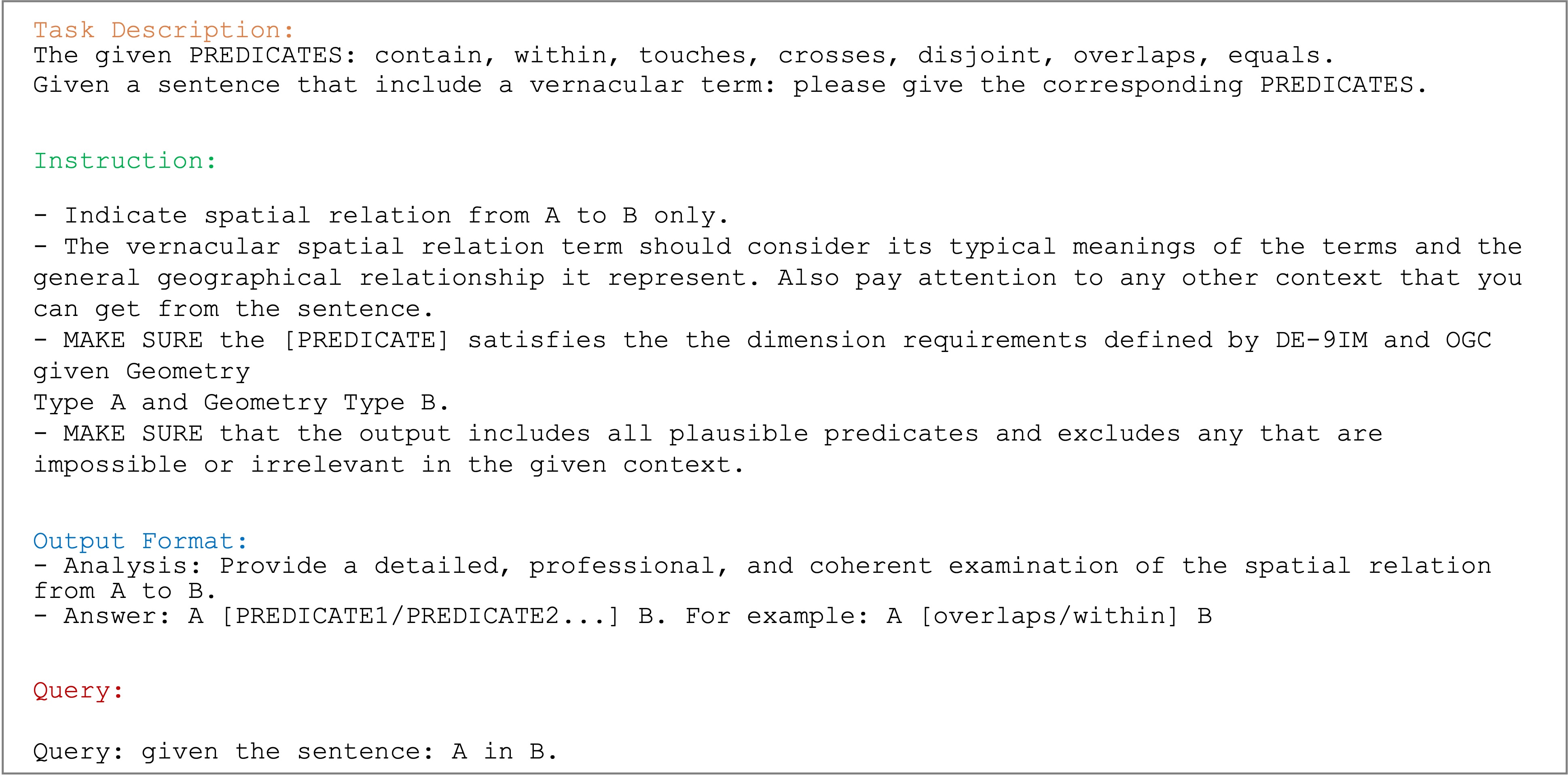}
    \caption{Prompt template used in the vernacular relation conversion task.}
    \label{fig:prompt_task_3}
\end{figure}

We adopt the workflow in Figure \ref{fig:workflow_3} to evaluate the capability of LLMs in task 3. The workflow begins with collecting geographic entities from a Web document knowledge database (DBpedia, structured knowledge based on Wikipedia), where named entity recognition is used to extract place names and vernacular spatial relation descriptions. These place names, such as ``UW-Madison" and ``Madison, Wisconsin" are then used to retrieve relevant geographic data and corresponding attributes from a spatial database to provide context such as geometry type and place type, with their topological spatial relations identified through GIS tools.
The collected spatial relations between two places are formatted as ``A \{vernacular topological relation\} B" (e.g., A is home to B) for evaluation, where A and B are symbolic placeholders representing two places. The specific locations in geometries are not disclosed, allowing for a generalized discussion of topological spatial relations without actual geographic context. The context will be provided at the end of the text input to support in-context reasoning. 
The context evaluated in our experiments is in Table \ref{tab:context_task_3}. 
The prompts are crafted with the template as shown in Figure \ref{fig:prompt_task_3} and fed into an LLM (e.g., GPT-4) to convert vernacular descriptions to topological spatial relations. We run the model multiple times to identify the possible converted topological predicates and the preference of an LLM. The output topological predicates are then compared with the ground truth predicates calculated by the GIS tool for evaluation. We also compare the performance when no contextual information is provided. This workflow allows us to evaluate the effectiveness of LLMs for analyzing informal topological relations between two entities and to assess the impact of contextual information on performance.

\begin{table}[h]
	\footnotesize
	\centering
	\caption{The textual description from DBPedia and topological predicate conversion examples.}
	\label{tab:context_task_3}
	\begin{tabular}{|c|c|c|l|}
		\hline
		Description & Context    & Predicate & \multicolumn{1}{c|}{Example}                                                                                \\ \hline
		bordered by & No context & touches   & \begin{tabular}[c]{@{}l@{}}Glendora is bordered by Azusa.\\ $\rightarrow$  A is bordered by B.\end{tabular} \\ \hline
		along &
		Geometry type &
		crosses &
		\begin{tabular}[c]{@{}l@{}}Luling is along the San Marcos River.\\ $\rightarrow$ A is along B.  A is Polygon, and B is LineString.\end{tabular} \\ \hline
		located on &
		Place type &
		crosses &
		\begin{tabular}[c]{@{}l@{}}Located on Interstate 10, Weimar is a small community.\\ $\rightarrow$  A is located on B. A is city, B is highway.\end{tabular} \\ \hline
		on the shore of &
		Place name &
		overlaps &
		\begin{tabular}[c]{@{}l@{}}Racine is located on the shore of Lake Michigan.\\ $\rightarrow$ A is on the shore of B. A is Racine in Wisconsin, B is lake Michigan.\end{tabular} \\ \hline
	\end{tabular}
\end{table}

The evaluation metrics for the vernacular relation conversion task are as follows. 

\begin{enumerate}[series=outerlist] 
    \item Frequency: The count of correctly returned predicates across all experiments.
    
    \item Accuracy: The ratio of the frequency of correctly returned predicates to the total number of generated outputs for each conversion pair.
\end{enumerate}

    

\begin{enumerate}[resume=outerlist] 
\item Entropy: The information entropy~\citep{shannon1948mathematical} of the returned predicates assesses the level of randomness in converting vernacular descriptions into topological predicates. Smaller entropy values indicate a higher likelihood of certain predicates being preferred over others. The metric is computed as:

$$
H = -\sum_{rel \in R} p_{rel} \log(p_{rel}),
$$

where $p_{rel}$ represents the probability of a specific topological predicate $rel$ appearing in the outputs for the given context-conditioned conversion pair.

\end{enumerate}

\section{Data and Experiments}\label{sec:experiments}
\subsection{Data processing}
\subsubsection{Extracting topological spatial relations from spatial database}

We construct real-world multi-sourced geospatial datasets for our study. The study area for \textbf{Task 1 and Task 2} is the city of Madison, Wisconsin, United States. The following datasets are collected.

\begin{itemize}
    \item OpenStreetMap road network data (including links and intersections) using \textit{OSMnx}. \footnote{http://osmnx.readthedocs.io/}

    \item Points of interest (POIs) categorized by \textit{SLIPO}. \footnote{http://slipo.eu/}
    
    \item Land parcels from \textit{Wisconsin Statewide Parcel Map Initiative}. \footnote{https://www.sco.wisc.edu/parcels/data/}

    \item Census block groups from \textit{U.S. Census Bureau}. \footnote{https://www2.census.gov/geo/tiger/TIGER2020PL/LAYER/BG/}  
    
\end{itemize}

Our evaluation tasks focus on the spatial objects with \textit{Point, LineString}, and \textit{Polygon} geometry types, assessing their topological spatial relations. All the computations are performed by using the \textit{GeoPandas} package in Python.

\textbf{Task 1 and Task 2} share the same dataset of triplets. For each combination of \{geometry type $A$, predicate, geometry type $B$\}, we obtain 200 triplets. Among these, 160 are allocated for training the random forest classification model, while the remaining 40 triplets are reserved for evaluation. Additionally, we set aside 25 extra triplets as candidate examples to facilitate few-shot learning. Due to the imbalanced distribution of topological spatial relations within the real-world dataset, we employ multiple strategies for sampling a sufficient number of triplets for fair comparisons:

\begin{enumerate}[series=outerlist]
    \item For topological spatial relations including ``within", ``contains", ``overlaps", ``touch" and ``crosses", we opt to select a subset of spatial objects and conduct spatial joins to obtain the required triplets.

    \item Regarding the ``equals" relationship. we manually created the equivalent spatial entities to preserve the topological spatial relations while making direct identification from geometry coordinate matching challenging.

        \begin{enumerate}[label=\alph*., series=innerlist]
            \item For \textit{Point}, include only points with identical coordinates.

            \item For \textit{LineString}, interpolate an additional 10\% of points along the lines, ensuring that the added points did not alter the original shape.

            \item For \textit{Polygon}, loop the origin point and interpolate additional points along the boundaries.
        \end{enumerate}
\end{enumerate}

\begin{enumerate}[resume=outerlist]
    \item We restrict the occurrence of ``disjoint" to cases where the subject geometry does not touch or overlap but lies within a smaller buffer of the objects (i.e., nearby entities), to avoid easy identification when two spatial entities are far apart, thus enhancing the evaluation on the differentiation of topological predicates.
\end{enumerate}

In \textbf{Task 2}, we further exclude the ``disjoint'' relations since most real-world geographic entities are disjoint from each other, yielding 40$\times$26=1040 triplets for retrieving the subject or object geographic entities.

\subsubsection{Topological spatial relations from DBpedia/Wikipedia}

For \textbf{Task 3}, we have gathered a total of 1078 unique triplets based on the recognized geographic entities from DBpedia/Wikipedia documents, which we combine with everyday descriptions of topological spatial relationships. We then utilize this extracted data to evaluate GPT-4's capabilities in task 3 as described in Section \ref{sec:vernacularrelations}.

Specifically, we downloaded and refined place descriptions within the States of Wisconsin, Texas, and California from the knowledge base \textit{DBpedia} \footnote{https://www.dbpedia.org/}, which is the linked data format of Wikipedia and has been previously used in place name disambiguation task~\citep{hu2014improving}. The data extraction and processing steps are structured as follows:

\begin{enumerate}[series=outerlist] 
\item Named entity recognition: From each administrative region's abstract ``dbo:abstract''), we extract all place names that can be found in OpenStreetMap, forming the basis for subsequent topological spatial relation identification.

\item Textual spatial relation extraction:  For each pair of place names within a DBpedia abstract, we use GPT-4 to extract topological spatial relation terms found between the entities in the text. When hierarchical place relationships are described, our approach only captures direct relations between a subject and each individual object, omitting implicit transitive relations among the objects themselves. For instance, from the sentence “a city A in a County B, State C,” we extract (A, in, B) but skip (A, in, C) and (B, in, C).

\item Manual verification: We manually review all the extracted spatial relation descriptions to ensure that they indicate topological relations and that the use of the two place names as subjects or objects in the sentence is semantically correct.

\item Description unification: The text descriptions on DBpedia are standardized for consistency. For example, phrases like ``is home to", ``home to" or ``home of",  are unified as ``is home to".

\item Context-conditioned conversion pairs extraction: We identify how vernacular descriptions depend on the following context to convert them to formal topological predicates.

\begin{enumerate}[label=\alph*., series=innerlist]
    \item Invariant to context: If a vernacular description consistently corresponds to the same topological predicate, we create a context-conditioned conversion pair (description, predicate, N/A).
\end{enumerate}

For descriptions that can be converted to multiple formal topological predicates, we associate them with specific contexts for one-on-one conversion.

\begin{enumerate}[label=\alph*.,resume=innerlist]
    \item Place types as context: If grouping by description, place type $A$ and place type $B$, results in a unique topological predicate, we create the pair (description, predicate, place type $A$/place type $B$). Place types are extracted from OpenStreetMap data tags.
    
    \item Geometry types as context: If grouping by description, geometry type $A$ and geometry type $B$, results in a unique predicate, we create the pair (description, predicate, geometry type $A$/geometry type $B$).

    \item Place names as context: Each pair of places can have a unique topological relationship. We create the pair (description, predicate, place name $A$/place name $B$), assuming the LLMs have some knowledge about place names.
    
\end{enumerate}

It is possible that more than one context can assist with one-on-one mapping from a vernacular description to a formal topological predicate. We may retain multiple contexts to compare their effectiveness. For example, the conversion between ``is bordered by" and ``touches" can be identified using place types (is bordered by, town/city, touches), geometry types (is bordered by, Polygon/MultiPolygon, touches), and place names (is bordered by, Aliso Viejo, California/Laguna Beach, California, touches).


\item Data filtering: Only frequently observed context-conditioned conversion pairs are retained for evaluation.

    \begin{enumerate}[label=\alph*., series=innerlist]
        \item  In the cases of invariant to context, place type as context, and geometry types as context, we retain pairs that occur at least 5 times for evaluation.

        \item In the case of place names as context, we first filter (description, predicate) that occur at least 5 times, and then sample 5 pairs for each combination.
    \end{enumerate}
    
\end{enumerate}

Among the 1078 records extracted from DBPedia abstracts of places in the states of Wisconsin, Texas, and California, 212 explicitly refer to directional and distance spatial relations and were thus removed as this research focused on topological relations. The analytical results of task 3 using the remaining records will be presented in Section \ref{sec:vernacularresult}.


\subsection{Experiment Models}
In this research, we perform evaluation tasks based on the following models:

\subsubsection{Embedding models}

We encode WKT geometries into embeddings and process spatial queries using ``text-embedding-ada-002" and ``text-embedding-3-large" provided by OpenAI\footnote{https://openai.com/blog/new-embedding-models-and-api-updates}, with output embedding dimensions of 1536 and 3072 respectively. 

\subsubsection{Reasoning models}

In our evaluation tasks, we employ GPT-3.5-turbo, GPT-4, and DeepSeek-R1-14B as the LLM-based reasoning models. While performance varies by task, these models have demonstrated potential in commonsense reasoning and in-context learning on certain benchmarks. GPT-3.5-turbo and GPT-4 are primarily optimized for few-shot learning, whereas DeepSeek-R1-14B emphasizes zero-shot capabilities and may experience a decline in performance when few-shot prompting is applied~\citep{guo2025deepseek}.

\subsubsection{Model settings}

\begin{enumerate}[] 

    \item Random Forest classifier: The number of estimators (trees) in the Task 1 classifier is set to 100. 
    
    \item Temperature settings for GPT-3.5-turbo and GPT-4: For the topological relation qualification task, we set the temperature to 0 to encourage more deterministic outputs. However, achieving full reproducibility remains challenging, even with a temperature of 0, as discussed by \citet{blackwell2024towards}. Conversely, generating synthetic geometries to support semantic search in Task 2, employs a higher temperature of 0.7 for greater creativity.

    \item Temperature settings for DeepSeek-R1-14B: The temperature of the topological relation qualification task is set to be 0.6 to better exploit the reasoning ability of DeepSeek, given its emphasis on deeper, more deliberate thinking.
\end{enumerate}

\section{Results} \label{sec:results}


\subsection{Topological spatial relation qualification}

\subsubsection{Validity of the output}

Before diving into the effectiveness of using LLMs to qualify spatial relationships, a validity check is necessary because of the inherent nondeterministic nature of generative AI models. Furthermore, beyond validating the output as a valid format of \{Geometry type $A$, predicate, Geometry type $B$\}, it is essential to focus on grounding the qualitative spatial reasoning in the matched geometry types and topological relations. 

The validity results of the output are shown in Table \ref{tab:t1_valid}. The random forest classifier using the LLM-generated embeddings consistently produced valid output on the test dataset. This highlights that the sentence embedding models can effectively preserve geometry types in the WKT format of geometries, aligning with previous research which encoded WKT by aggregating the token embeddings from GPT-2 and BERT ~\citep{ji2023evaluating}. While GPT-4 and GPT-3.5-turbo largely adhere to the instructions in the desired format, even with the CoT generation, it is more challenging for DeepSeek-R1-14B to strictly output the desired format (but still achieved over 0.9 validity accuracy). When tested with few-shot prompting, the DeepSeek-R1-14B largely ignored the provided examples and adhered to its typical reasoning patterns. As a result, we did not include these results in our evaluation. The highest validity of GPT-4 model suggests that a language model that is characterized by a larger number of parameters, broader training data, and stronger alignment with human instructions, may also possess a better understanding of the definitions of the DE-9IM topological predicates.

\begin{table}[ht]
	\footnotesize
	\centering
	\caption{The validity accuracy of the outputs (N/A: not available).}
	\begin{tabular}{|c|c|c|c|c|c|}
		\hline
		Approach & LLM                    & Prompt        & Format & Geometry type & Predicate \\ \hline
		\multirow{2}{*}{\begin{tabular}[c]{@{}c@{}}Random\\ Forest\end{tabular}} &
		\begin{tabular}[c]{@{}c@{}}text-embedding\\ -ada-002\end{tabular} &
		N/A &
		1 &
		1 &
		1 \\ \cline{2-6} 
		&
		\begin{tabular}[c]{@{}c@{}}text-embedding\\ -3-large\end{tabular} &
		N/A &
		1 &
		1 &
		1 \\ \hline
		\multirow{12}{*}{\begin{tabular}[c]{@{}c@{}}Question\\ answering\end{tabular}} &
		\multirow{5}{*}{GPT-3.5-turbo} & Zero-shot & 0.959 & 1 & 0.911 \\ \cline{3-6} 
		&                        & Zero-shot-dim & 0.999  & 1             & 0.927     \\ \cline{3-6} 
		&                        & Few-shot      & 1      & 1             & 0.901     \\ \cline{3-6} 
		&                        & Zero-shot-CoT & 0.944  & 1             & 0.944     \\ \cline{3-6} 
		&                        & Few-shot-CoT  & 0.998  & 1             & 0.894     \\ \cline{2-6} 
		& \multirow{5}{*}{GPT-4} & Zero-shot     & 1      & 0.996         & 0.997     \\ \cline{3-6} 
		&                        & Zero-shot-dim & 1      & 0.999         & 0.999     \\ \cline{3-6} 
		&                        & Few-shot      & 1      & 0.999         & 0.992     \\ \cline{3-6} 
		&                        & Zero-shot-CoT & 0.984  & 0.990          & 0.968     \\ \cline{3-6} 
		&                        & Few-shot-CoT  & 1      & 0.999         & 0.999     \\ \cline{2-6} 
		& \multirow{2}{*}{DeepSeek-R1-14B} & Zero-shot     & 0.936  &  0.996         & 0.913   \\ \cline{3-6} 
		&                       & Zero-shot-dim & 0.919  & 0.998         & 0.913    \\ \hline
	\end{tabular}
	\label{tab:t1_valid}
\end{table}


\subsubsection{Classification metrics}

Table \ref{tab:t1_metrics} presents the results of the topological predicate classification task. Both the embedding-based random forest and geospatial question-answering with GPT models can achieve an accuracy of over 0.6. This suggests that identifying topological spatial relationships from the WKT format of geometries with LLMs is promising but remains challenging. Failure to recover the topological spatial relations from embeddings suggests a potential information loss through text tokenization. Incorrectly classified topological relations often cluster within the conceptual neighborhoods or resemble each other (with a small distance), while confusion may also arise from the diverse semantics of topological spatial predicates.

\begin{table}[h]
	\footnotesize
	\centering
	\caption{The topological predicate classification metrics.}
	\begin{tabular}{|c|c|c|c|c|}
		\hline
		Approach                                                                       & LLM                                                               & Prompt        & Accuracy & Dist(Incorrect)  \\ \hline
		\multirow{2}{*}{\begin{tabular}[c]{@{}c@{}}Random\\ Forest\end{tabular}}       & \begin{tabular}[c]{@{}c@{}}text-embedding\\ -ada-002\end{tabular} &             N/A  & 0.633    & 1.449 \\ \cline{2-5} & \begin{tabular}[c]{@{}c@{}}text-embedding\\ -3-large\end{tabular} &     N/A      & 0.632    & 1.419 \\ \hline
		\multirow{12}{*}{\begin{tabular}[c]{@{}c@{}}Question\\ answering\end{tabular}} & \multirow{5}{*}{GPT-3.5-turbo}                                    & Zero-shot     & 0.423    & 1.331 \\ \cline{3-5}           &                                                                   & Zero-shot-dim & 0.408    & 1.360 \\ \cline{3-5}           &                                                                   & Few-shot      & 0.479    & 1.595 \\ \cline{3-5}           &                                                                   & Zero-shot-CoT & 0.443    & 1.370 \\ \cline{3-5}           &      & Few-shot-CoT  & 0.465    & 1.174 \\ \cline{2-5} 
		& \multirow{5}{*}{GPT-4}                                            & Zero-shot     & 0.632    & 1.238 \\ \cline{3-5}  
		&  		& Zero-shot-dim & 0.635    & 1.212 \\ \cline{3-5} 
		& 		 & Few-shot      & 0.666    & 1.272 \\ \cline{3-5} 
		&                                                                   & Zero-shot-CoT & 0.610    & 1.256 \\ \cline{3-5} 
		&                                                                   & Few-shot-CoT  & 0.627    & 1.225 \\ \cline{2-5} 
		& \multirow{2}{*}{DeepSeek-R1-14B}                     & Zero-shot     &  0.534    & 1.257   \\ \cline{3-5} 
		&    &Zero-shot-dim  & 0.557    & 1.260   \\ \hline
	\end{tabular}
	\label{tab:t1_metrics}
\end{table}

For GPT-3.5-turbo and GPT-4, among the four types of prompts (introduced in Section~\ref{sec:task1}), few-shot learning achieved the best performance with pairs of geometries and their topological relationships for LLMs to learn in context. GPT-4 with few-short promoting achieved 0.66 accuracy. The findings highlight the importance of prompt engineering in the use of LLMs and the critical role of understanding spatial contexts in improving geospatial query processing accuracy and reliability.
However, chain-of-thought (CoT) prompts, which have demonstrated improvement in many other tasks~\citep{wei2022chain}, did not yield the expected benefit in our spatial reasoning evaluation experiments. As mentioned by \cite{yang2024thinking}, CoT reasoning can sometimes induce unreliable or counterproductive outputs in spatial reasoning tasks. Upon analyzing the generated rationale, we observed that when LLMs are prompted with  ``Let's think step by step", they attempt to check the topological spatial predicates one by one based on their respective definitions from the OGC standard. Few-shot-CoT prompts, on the other hand, were explicitly designed with examples grounded in scientific definitions and logical decision processes proposed in \cite{clementini1993small}, aiming to ``teach" the models to reason about topological spatial relations from analysis on interiors, boundaries, and exteriors. Despite this structured approach, the accuracy declined due to cascading errors in intermediate steps, such as failing to determine whether the interiors of two geometries intersect at the very beginning. With an explicit reasoning process, DeepSeek-R1-14B outperformed GPT-3.5-turbo (20B parameters). Analysis of its thought generation reveals that, rather than always iteratively checking candidate answers, the model often employed more intuitive reasoning strategies, such as mental mapping (e.g., ``Let me plot them mentally'') and self-verification (e.g., ``In WKT, a LINESTRING is just a sequence of points connected by straight lines. If it starts and ends at the same point, it doesn't automatically become a polygon''). Although these reasoning patterns may appear convincing when interpreting individual geometries, they often fall short when reasoning about spatial relations between two geometries. This is mainly due to an overreliance on superficial, linear interpretations of coordinate information, rather than a holistic understanding of topological spatial relationships across the plane.

\subsection{Spatial query processing}

Based on the superior performance in task 1, the experiments in task 2 only used GPT-4 as the geometry generator and text-embedding-3-large as the embedding model.

\subsubsection{Direct query}

We first identified an effective query format for geospatial semantic search~\citep{hu2015metadata}, which is the foundation for applying query expansion in understanding geospatial semantics. As shown in Table \ref{tab:t2_performance}, specifying the subject geometry type would achieved higher performance due to a narrowed mapping space to the same geometry type. In the following experiments, we assumed that the user query with the geometry type (e.g. retrieving a street from the spatial database implies LineString), and further investigated the factors that may impact the effectiveness of query expansion using LLMs. 

%
%
%

\begin{table}[ht]
	\footnotesize
	\centering
	\caption{Spatial query performance comparison results.}
	\begin{tabular}{|c|cc|c|c|c|c|}
		\hline
		Target &
		\multicolumn{2}{c|}{Query Format} &
		MRR &
		Hits@5 &
		Hits@10 &
		Hits@20 \\ \hline
		\multirow{4}{*}{Subject} &
		\multicolumn{1}{c|}{\multirow{2}{*}{Direct query}} &
		Abstract as ``geometry" &
		0.081 &
		0.131 &
		0.161 &
		0.194 \\ \cline{3-7} 
		&
		\multicolumn{1}{c|}{} &
		Specify the subject geometry type &
		0.152 &
		0.212 &
		0.26 &
		0.29 \\ \cline{2-7} 
		&
		\multicolumn{1}{c|}{\multirow{2}{*}{Expanded query}} &
		\begin{tabular}[c]{@{}c@{}}Direct query + \\ one LLM-generated geometry\end{tabular} &
		0.18 &
		0.238 &
		0.278 &
		0.328 \\ \cline{3-7} 
		&
		\multicolumn{1}{c|}{} &
		\begin{tabular}[c]{@{}c@{}}Direct query + \\ three LLM-generated geometry\end{tabular} &
		0.169 &
		0.232 &
		0.28 &
		0.32 \\ \hline
		\multirow{4}{*}{Object} &
		\multicolumn{1}{c|}{\multirow{2}{*}{Direct query}} &
		Original predicate &
		0.105 &
		0.131 &
		0.17 &
		0.211 \\ \cline{3-7} 
		&
		\multicolumn{1}{c|}{} &
		Reversed predicate &
		0.152 &
		0.219 &
		0.256 &
		0.297 \\ \cline{2-7} 
		&
		\multicolumn{1}{c|}{\multirow{2}{*}{Expanded query}} &
		\begin{tabular}[c]{@{}c@{}}Original predicate + \\ one LLM-generated geometry\end{tabular} &
		0.15 &
		0.215 &
		0.261 &
		0.302 \\ \cline{3-7} 
		&
		\multicolumn{1}{c|}{} &
		\begin{tabular}[c]{@{}c@{}}Reversed predicate + \\ one LLM-generated geometry\end{tabular} &
		0.179 &
		0.248 &
		0.294 &
		0.333 \\ \hline
	\end{tabular}
	\label{tab:t2_performance}
\end{table}

\subsubsection{Synthetic geometry generation}

\begin{figure}
    \centering
    \includegraphics[width=.8\linewidth]{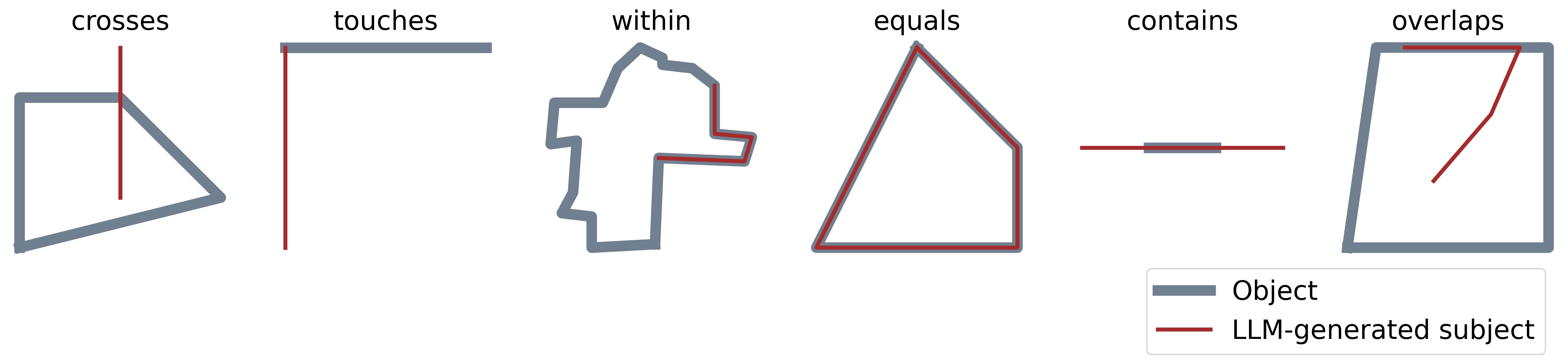}
    \caption{Synthetic geometries generated by GPT-4 for LineString/LineString relations}
    \label{fig:t2_relations}
\end{figure}

An effective LLM-generated geometry is expected to maintain the same topological spatial relation with the given object as the subject entity while being close to the subject entity in the embedding space. 
Table \ref{tab:t2_geom} compares the validity of the LLM-generated geometries produced by different prompting approaches.  We find that 1) GPT-4 effectively comprehends spatial queries and generates geometries in a valid WKT format, and 2) GPT-4 demonstrates a notable level of spatial reasoning regarding the reference object, even in the zero-shot setting, as indicated by the high relation-preserving accuracy (over 0.72) and the low topological distance in the conceptual neighborhood graph~\citep{reis2008conceptual}. Figure \ref{fig:t2_relations} presents examples of the LLM-generated geometries generated by GPT-4. In the following section, we will check the usefulness of such synthetic geometries generated by zero-shot prompts in enriching spatial query processing.
%

\begin{table}[h]
	\footnotesize
	\centering
	\caption{Validity of LLM-generated geometries using different prompts.}
	\begin{tabular}{|c|c|c|c|c|}
		\hline
		Prompt            & Valid\_WKT & Geometry Type & Predicate & Topological Distance \\ \hline
		Zero-shot         & 0.999      & 1             & 0.763     & 1.142 \\ \hline
		Zero-shot-Check   & 0.998      & 1             & 0.755     & 1.075 \\ \hline
		Few-shot          & 0.996      & 1             & 0.728     & 1.177 \\ \hline
		Few-shot-Negative & 0.997      & 1             & 0.754     & 1.212 \\ \hline
	\end{tabular}
	\label{tab:t2_geom}
\end{table}

\subsubsection{Expanded query on subject retrieval }
As shown in Table \ref{tab:t2_performance}, retrieving a subject based on the embeddings encoded from the expanded spatial queries remains challenging. However, including an LLM-generated geometry enhanced the probability of ranking the target subject higher among all candidates. Over 23\% of the subjects were ranked within the top 5 candidates. But adding additional synthetic geometries did not appear to provide further improvements.

\subsubsection{Performance comparison on object retrieval}

While the above experiment primarily focuses on retrieving the subject in a triplet, we proceed to evaluate the performance of object retrieval. For a given triplet, we tested the queries formulated with either the original predicate describing the spatial relationship from the subject to the object (e.g., ``Retrieve a Point which A contains''), or the reversed predicate referring to the subject (e.g., ``Retrieve a Point which is within A''). The results of object retrieval are also summarized in Table \ref{tab:t2_performance}. 

Among the object-retrieval query formats, the queries with the original predicate that maintained the subject-to-object directionality yielded worse performance. When we manually reversed the topological spatial relation and treated the object as the subject, the performance matched its counterpart for subject retrieval in Table \ref{tab:t2_performance}, highlighting the importance of structuring spatial queries to align with everyday language patterns commonly used for spatial reference.

\subsection{Vernacular topological relation conversion}\label{sec:vernacularresult}

In Task 3, we collected textual descriptions of topological spatial relations between two places and attempted to identify mapping patterns between these descriptions and the corresponding context-conditioned topological spatial relations. These mappings were then used as input to GPT-4 (giving its superior performance in previous tasks) to evaluate their ability to convert textual descriptions into formal topological spatial relations.

\subsubsection{Conversion pairs invariant to context}

Table \ref{tab:t3_invariant} lists the six descriptions that consistently map to the same topological relationship in our dataset. However, the results show varying levels of conversion accuracy from vernacular descriptions to preferred formal topological predicates. While the ground truth topological relationships were likely to be implied from ``share border with'' and ``is the location of'', ``is an enclave of'' was  interpreted as \textit{within} instead of \textit{touches} or \textit{disjoint}. Even though GPT-4 could infer an \textit{overlaps} relation from ``has part of the population in" in all ten experiments, the model might be unsure about its answer and would provide multiple topological predicate alternatives. Despite the subtle difference between ``midway" and ``halfway", a higher entropy of ``halfway" indicates greater randomness in the conversion. 


\begin{table}[h]
	\footnotesize
	\centering
	\caption{Result of topological relation conversion pairs invariant to context.}
	\label{tab:t3_invariant}
	\begin{tabular}{|c|c|c|c|c|}
		\hline
		Description                   & Predicate & Frequency & Accuracy & Entropy \\ \hline
		share border with             & touches   & 10        & 1.000    & 0.000   \\ \hline
		has part of the population in & overlaps  & 10        & 0.588    & 0.435   \\ \hline
		is the location of            & contains  & 9         & 0.818    & 0.244   \\ \hline
		is midway between C and       & disjoint  & 6         & 0.600    & 0.560   \\ \hline
		is halfway between C and      & disjoint  & 6         & 0.500    & 0.817   \\ \hline
		is an enclave of              & touches   & 1         & 0.100    & 0.167   \\ \hline
	\end{tabular}
\end{table}

\subsubsection{Conversion pairs conditioned on place types or geometry types}

The comparisons between scenarios with and without place-type/geometry-type context are illustrated in Table \ref{tab:place_type} and Table \ref{tab:geometry_type} respectively. Our initial hypothesis was that including contextual information in the prompt would reduce ambiguity, resulting in a higher frequency of correct predicate predictions, improved accuracy, and lower entropy using LLMs. However, these improvements were highly instance-dependent and not consistently observed across all the conversion pairs evaluated in our experiments, indicating GPT-4's limitation in considering all possible interpretations of the given vernacular description. This limitation was also evident in pairs with 0 accuracy, where the model consistently outputted the same incorrect answer. In other instances, GPT-4 still struggled to determine the appropriate topological predicates for certain vernacular descriptions.

\begin{table}[ht]
	\footnotesize
	\centering
	\caption{Result of topological relation conversions using place type as context}
	\resizebox{\textwidth}{!}{%
		\begin{tabular}{@{}llcccccc@{}}
			\toprule
			\textbf{Description}        & \textbf{Predicate} & \textbf{Spatial Context} & \textbf{Frequency} & \textbf{Accuracy} & \textbf{\parbox{2cm}{Accuracy\\without Context}} & \textbf{Entropy} & \textbf{\parbox{2cm}{Entropy\\without Context}} \\ \midrule
			is home to                  & contains           & city/amenity             & 10                 & 1                 & 1                                  & 0               & 0                               \\
			borders                     & touches            & city/municipality        & 10                 & 1                 & 1                                  & 0               & 0                               \\
			is located in               & within             & town/county              & 10                 & 1                 & 1                                  & 0               & 0                               \\
			is located in               & within             & city/state               & 10                 & 1                 & 1                                  & 0               & 0                               \\
			is bordered by              & touches            & town/city                & 10                 & 1                 & 1                                  & 0               & 0                               \\
			is adjacent to              & touches            & city/municipality        & 10                 & 1                 & 0.909                              & 0               & 0.157                           \\
			borders                     & touches            & city/city                & 10                 & 1                 & 1                                  & 0               & 0                               \\
			is in                       & within             & village/county           & 10                 & 0.909             & 0.909                              & 0.157           & 0.157                           \\
			is located in               & within             & village/county           & 10                 & 0.909             & 1                                  & 0.157           & 0                               \\
			is partly in                & overlaps           & city/county              & 10                 & 0.833             & 1                                  & 0.232           & 0                               \\
			is bounded by               & touches            & city/city                & 7                  & 0.7               & 0.333                              & 0.314           & 0.327                           \\
			connect C and               & crosses            & industrial/city          & 8                  & 0.4               & 0.474                              & 0.52            & 0.355                           \\
			extend into                 & overlaps           & city/county              & 5                  & 0.357             & 0.421                              & 0.561           & 0.491                           \\
			is surrounded by            & touches            & city/city                & 2                  & 0.167             & 0                                  & 0.232           & 0                               \\
			is between C and            & touches            & town/town                & 0                  & 0                 & 0.3                                & 0.211           & 0.773                           \\
			is surrounded by            & touches            & town/city                & 0                  & 0                 & 0                                  & 0               & 0                               \\
			is within                   & touches            & city/municipality        & 0                  & 0                 & 0                                  & 0               & 0                               \\ \bottomrule
		\end{tabular}%
	}
	\label{tab:place_type}
\end{table}

\begin{table}[ht]
	\centering
	\caption{Result of topological relation conversions using geometry type as context}
	\resizebox{\textwidth}{!}{%
		\begin{tabular}{@{}llcccccc@{}}
			\toprule
			\textbf{Description}        & \textbf{Predicate} & \textbf{Spatial Context} & \textbf{Frequency} & \textbf{Accuracy} & \textbf{\parbox{2cm}{Accuracy\\without Context}} & \textbf{Entropy} & \textbf{\parbox{2cm}{Entropy\\without Context}} \\ \midrule
			is in                       & within             & Polygon/MultiPolygon          & 10                 & 1                 & 0.909                                              & 0               & 0.157                                              \\
			is neighboring              & touches            & Polygon/Polygon               & 10                 & 1                 & 1                                                  & 0               & 0                                                  \\
			is bordered by              & touches            & Polygon/MultiPolygon          & 10                 & 1                 & 1                                                  & 0               & 0                                                  \\
			is the county seat of       & within             & Polygon/Polygon               & 10                 & 1                 & 1                                                  & 0               & 0                                                  \\
			extend into                 & overlaps           & Polygon/Polygon               & 8                  & 0.714             & 0.421                                              & 0.307           & 0.491                                              \\
			connect C and               & crosses            & LineString/MultiPolygon       & 9                  & 0.412             & 0.474                                              & 0.348           & 0.355                                              \\
			is surrounded by            & touches            & Polygon/MultiPolygon          & 0                  & 0                 & 0                                                  & 0               & 0                                                  \\
			is on                       & crosses            & Polygon/LineString            & 0                  & 0                 & 0                                                  & 0               & 0.327                                              \\ \bottomrule
		\end{tabular}%
	}
	\label{tab:geometry_type}
\end{table}

\subsubsection{Conversion pairs with place names}

The accuracy and entropy of the topological relation conversions with place names were also compared to the metrics obtained without the context. As shown in Table \ref{tab:t3_name}, mentioning place names did not necessarily improve the accuracy of the conversion or guide the LLM to a preferred answer. GPT-4's explanation indicates that 1) It focuses on the topological relationships between general geographic locations or boundaries rather than leveraging specific knowledge about each place; 2) The approach tends to exclude predicates possibly with inaccurate and abstract geometries. For instance, in analysis ``A is along B" when A is Brazos Bend, Texas, and B is Brazos River, the reasoning begins with ``This suggests a specific geographical relationship between a place (A) and a river (B). The term `along' typically indicates that A is situated in a linear arrangement adjacent to B, but not necessarily crossing it or being contained within it. ''

\begin{table}[]
	\footnotesize
	\centering
	\caption{Accuracy and Entropy changes for conversion pairs with place names}
	\begin{tabular}{|c|ll|}
		\hline
		Accuracy & \multicolumn{2}{c|}{\begin{tabular}[c]{@{}c@{}}Topological relation conversion pairs \\ (Order by the absolute values in change, $^*$ with entropy reduction)\end{tabular}} \\ \hline
		Improves &
		\begin{tabular}[c]{@{}l@{}}1) is bounded by$\rightarrow$touches$^*$ \\ 3) is suburb of$\rightarrow$touches\\ 5) is part of$\rightarrow$within$^*$ \\ 7) is partly in$\rightarrow$touches \\ 9) is between C and$\rightarrow$disjoint$^*$\end{tabular} &
		\begin{tabular}[c]{@{}l@{}}2) is surrounded by$\rightarrow$touches\\ 4) is on$\rightarrow$crosses\\ 6) is between C and$\rightarrow$touches$^*$\\ 8) is suburb of$\rightarrow$disjoint\\ 10) is near$\rightarrow$touches\end{tabular} \\ \hline
		Unchanged &
		\begin{tabular}[c]{@{}l@{}}Remains 1:\\ 1) includes$\rightarrow$contains \\ 3) is bordered by$\rightarrow$touches\\ 5) is located in$\rightarrow$within\\ Remains 0:\\ 7) is bordered by$\rightarrow$disjoint \\ 9) on the shore of$\rightarrow$overlaps\end{tabular} &
		\begin{tabular}[c]{@{}l@{}}\\2) borders$\rightarrow$touches\\ 4) is neighboring$\rightarrow$touches\\ 6) is the county seat of$\rightarrow$within\\   \\ 8) is in$\rightarrow$overlaps$^*$\\ 10) is within$\rightarrow$touches\end{tabular} \\ \hline
		Declines &
		\begin{tabular}[c]{@{}l@{}}1) is mostly in$\rightarrow$overlaps \\ 3) is along$\rightarrow$crosses$^*$\\ 5) is situated on$\rightarrow$overlaps\\ 7) extend into$\rightarrow$overlaps\\ 9) is home to$\rightarrow$contains\end{tabular} &
		\begin{tabular}[c]{@{}l@{}}2) is near$\rightarrow$disjoint\\ 4) is partly in$\rightarrow$overlaps\\ 6) is adjacent to$\rightarrow$touches\\ 8) is in$\rightarrow$within\\ 10) connect C and$\rightarrow$crosses$^*$\end{tabular} \\ \hline
	\end{tabular}
	\label{tab:t3_name}
\end{table}

\section{Discussion} \label{sec:discussion}

In this section, we would like to further discuss whether the confusion between the topological predicates aligns with the corresponding conceptual neighborhood of topological spatial relations~\citep{formica2018approximate, egenhofer1995modelling, egenhoferReasoningGradualChanges1992}, the confusion in geometry generation, and the confusion in vernacular description conversion.

\subsection{Confusion between topological predicates in topological spatial relation qualification}

\begin{figure}[!h]
    \centering
\includegraphics[width=0.9\linewidth]{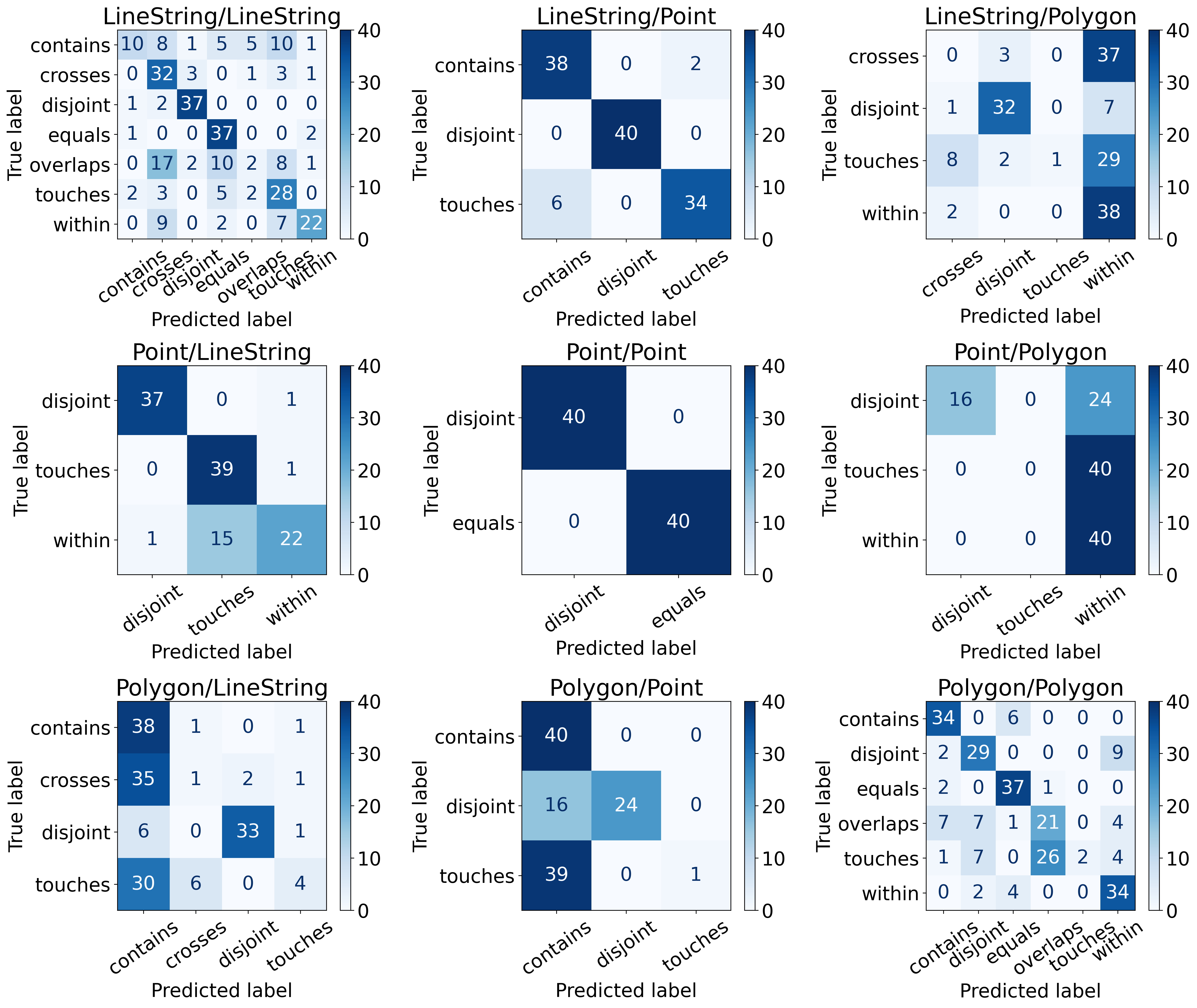}
    \caption{Confusion matrices between topological predicates in relation qualification.}
    \label{fig:t1_confusion} 
\end{figure}

When using GPT-4 (zero-shot learning) for topological spatial relation qualification, the confusion matrices for all the geometry type combinations are drawn in Figure \ref{fig:t1_confusion}. We compare the topological predicate pairs that may confuse GPT models with the classic conceptual neighborhood graphs in Figure \ref{fig:conceptual_neighborhood}. 
The observations are twofold: 
1) The most frequently confused topological spatial relation for a given predicate depends on the geometry types involved. 
For example, consider the predicate ``overlaps." In a Linestring/Linestring relationship, it is rarely classified correctly and is often confused with ``crosses", ``equals" or ``touches". However, in a Polygon/Polygon relationship, ``overlaps" is more likely to be correctly identified, though it may occasionally be confused with ``contains" or ``disjoint".
Another illustrative example involves the predicate ``touches." A Point ``touches" a Linestring or Polygon, or a Linestring ``touches" a Polygon is frequently mistaken as "within" while such confusion is less between two geometries of the same dimension, such as two Polygons or two Linestrings.
These examples suggest the varied degree to which an LLM understands formal geometry boundaries associated with geometry types, particularly their dimensions, which is crucial in identifying formal topological spatial relations. However, the constraint of formal definitions may contradict common conceptual interpretations, such as excluding a polygon from containing its boundary, leading to fewer occurrences in GPT-4's response.
2) For the same geometry type combination, distinguishing certain pairs of topological spatial relations is more challenging than others. These pairs mostly fall within the conceptual neighborhood, though exceptions exist. Take ``Linestring/Linestring" as an example. GPT-4 can identify ``crosses", ``disjoint", and ``equals" more accurately. However, it struggles with predicates like ``contains" and ``overlaps," frequently confusing them with ``crosses" or ``touches". The four topological spatial relations all require that the two geometries share elements like points or line segments and might be interchangeable in daily use. This challenge highlights that the ambiguous semantics of these predicates can encompass scenarios broader than their strict formal definitions. 
Overall, the issues observed in Task 1 actually reflect an alignment with everyday spatial reasoning. While formal definitions are precise and dimension-contingent, everyday language and intuitive reasoning often blur the distinctions.



\subsection{Confusion between topological predicates in geometry generation}

\begin{figure}[!h]
    \centering
    \includegraphics[width=0.9\linewidth]{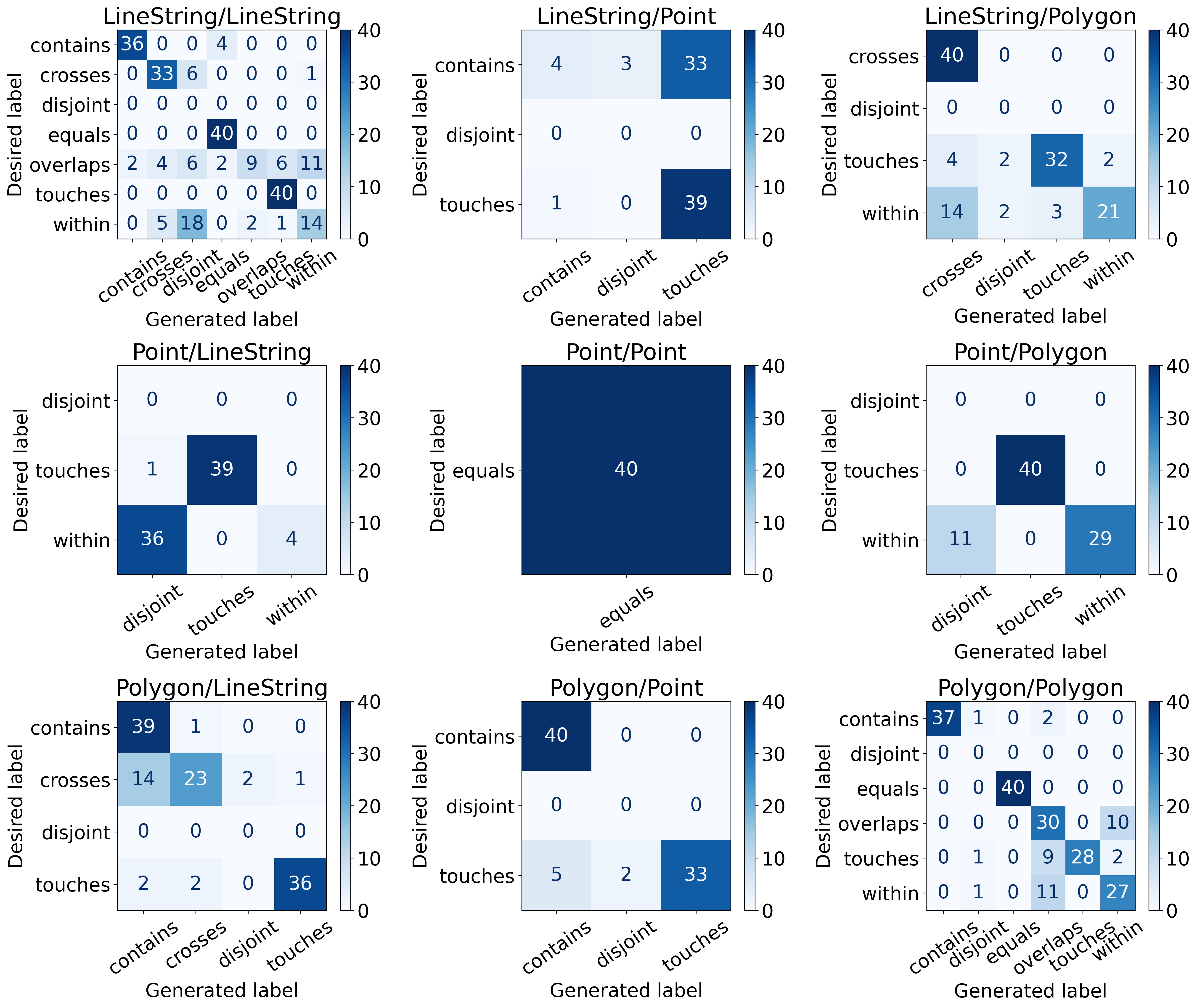}
    \caption{Confusion matrices between spatial predicates in geometry generation.}
    \label{fig:t2_confusion} 
\end{figure}

\begin{figure}[!h]
    \centering
    \includegraphics[width=1\linewidth]{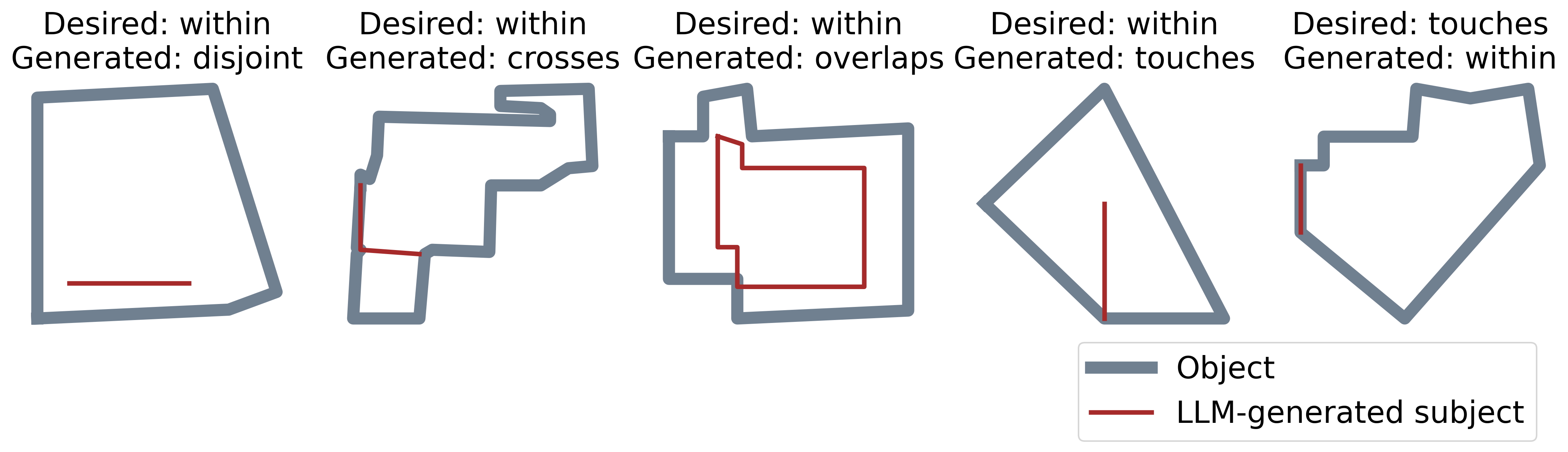}
    \caption{Invalid synthetic geometries generated by GPT-4 for LineString/LineString relations with close-shape objects.}
    \label{fig:t2_relations_wrong}
\end{figure}

The confusion pattern changes when leveraging an LLM to generate geometries given a spatial query and the required geometry type, as shown in Figure \ref{fig:t2_confusion}. 
The findings can be summarized as follows: 
1) Directionality in describing the topological relation between two geometry types matters. For example, generating a Polygon that “crosses” a LineString proves challenging for GPT-4, while the reversed query—generating a LineString that ``crosses” a Polygon—is handled more effectively. Similarly, the model is more successful in generating a Polygon that ``contains” a Point, LineString, or another Polygon but struggles to produce a Point, LineString, or Polygon that is ``within” a Polygon.
This asymmetry can be attributed to the model’s approach of extracting coordinates from the query geometry to construct the second geometry. This reliance limits the model’s ability to conceptualize spatial relationships beyond the provided coordinates. 
2) The geometry type of the reference object affects the results. Figure \ref{fig:t2_relations_wrong} provides examples of LineString/LineString topological spatial relations where the generated topological spatial relations were different from the predicate in the spatial query. As observed from these examples, even if the spatial queries specify the reference object geometry type as LineString, the model sometimes applies definitions for Polygons when a line forms a closed shape. In this case, when we manually changed the reference object geometry type into Polygon and recompute its topological relations with the LLM-generated geometry, 223 out of 391 queries (across all prompts) with closed geometries were found to exhibit the desired topological relationship. This observation suggests GPT-4's perception based on the provided coordinates over the geometry type specified in the text, inspiring us to further explore the cognition potential of the LLMs.

\subsection{Confusion between topological spatial relations in conversion}


While GPT-4 can accurately convert several vernacular descriptions to corresponding formal predicates, there are instances where its performance falls short. This can be attributed to the mismatch between crispy geometry and the vague human perception of place boundaries. We can further divide it into three categories. 
1) The abstraction of spatial entities' shapes in the spatial dataset may differ from those used in descriptions. For example, when converting ``is along" for Brazos Bend, Texas, and Brazos River, Texas, GPT-4 returned ``touches" when considering Brazos Bend as a region (Polygon) and ``within" when considering Brazos Bend as a Point. 
2) The computed topological spatial relations can be sensitive to the marked shape points, while human perception can tolerate such systematic errors, yielding the description of the relations in the conceptual neighborhood of the ground truth. A typical example is when the ground truth label of ``is suburb of" is ``disjoint", but the two cities look like they ``touch" each other on the map. 
3) Official geographic boundaries might differ from people's perception of a place~\citep{gao2017data}. In our dataset, ``is within" and ``is an enclave of" can map to ``touches". But the LLM would constantly output ``within". For instance, the City of Shullsburg, Wisconsin, and the Town of Shullsburg, Wisconsin, illustrate this discrepancy \footnote{\url{https://en.wikipedia.org/wiki/Shullsburg_(town),_Wisconsin}}. Although the City of Shullsburg is enclosed by the Town of Shullsburg, the city boundary is separated from the town boundary, creating a hole in the town boundary.
In summary, even though GPT-4's responses can be partly interpreted from the conceptual neighborhood of topological spatial relations, challenges remain due to the vagueness of real-world geographic entity boundaries and human perception of shapes and places.

\section{Conclusion and future work} \label{sec:conclusion}

This study focuses on the evaluation of the ability of LLMs including GPT-3.5, GPT-4, and DeepSeek-R1-14B to process, represent, and reason with topological spatial relations. 
Consequently, we designed a workflow to assess the efficacy of LLMs in addressing three typical problems on topological spatial relations. The core idea involves converting geometric objects into textual strings (WKT), which can then be decoded and utilized for spatial reasoning. The first task, topological spatial relation qualification, focuses on determining if such textual representation can retain the necessary geometric information for deriving named topological predicates. The second task explores the feasibility of conducting geospatial queries through semantic search, where LLMs can generate a geometry to augment the query and also generate embeddings. The third task presents an everyday scenario where an LLM serves as a translator to convert vernacular descriptions of spatial relations into formalized topological predicates based on their capability to understand linguistic patterns. 

From the multi-source geospatial datasets, we extract triplets to represent topological spatial relations in real-world spatial configurations. Using the triplets as input, we have compared the performance on the three evaluation tasks with ground truth data. 
In Task 1, both the random forest and GPT-based reasoning models can identify most relations correctly (over 0.6 accuracy on average), while some relations can be confounding. For GPT-3.5-turbo and GPT-4, few-shot prompt engineering is essential to improve the performance while CoT prompting strategy had a negative impact on our topological spatial relation inference task. The thought generation process and the self-verification allow DeepSeek-R1-14B to perform spatial reasoning more intuitively and outperformed GPT-3.5-turbo in accuracy. Further comparison with the conceptual neighborhood allows for a more quantitative understanding of the errors. 
Even though task 2 further verifies the challenge of replacing spatial queries with semantic search. However, improvements can be observed when we customize the query and augment it with LLM-generated geometries. The LLM-generated geometries are not only valid WKT but also have high accuracy (up to 0.76) in preserving topological spatial relations (or within their conceptual neighbors).  
In task 3, the improvement of LLMs to reduce ambiguity in spatial queries is relatively limited. However, in most cases, the generated outputs fall into the conceptual neighborhood of the ground-truth topological predicate. Moreover, given various contexts, the changes in the preferred response show the ability of the LLM to reason about it using commonsense knowledge and the typical spatial configurations. Interestingly, adding the geometry-type context in prompts has derived more performance improvement compared to the cases with adding the place-type context, but the performance of adding context or without context varies by instance.

In conclusion, through the three tasks and intensive experiments, we systematically approach the overarching question of LLM's ability in understanding geometry information and their topological spatial relations, moving from the broader challenge to more targeted strategies involving spatial context, tailored prompting techniques, and specialized domain knowledge in GIScience.

However, it is essential to acknowledge the limitations of our work. First, our focus was primarily on in-context learning, and we did not explore fine-tuning approaches, which could potentially yield further performance improvements. Retrieval-Augmented Generation (RAG)~\citep{lewis2020retrieval} presents a promising approach for enhancing the qualitative spatial reasoning capabilities LLMs by integrating external spatial databases, GIS tools and domain-specific knowledge from GIScience. Unlike in-context learning, which allows for intuitive qualitative spatial reasoning, the effective implementation of RAG relies on the precise generation of formalism-based spatial queries from natural language input along with reliable high-resolution datasets. Improving the translation from natural language to into symbolic form and logic also opens the door to neurosymbolic approaches~\citep{sheth2023neurosymbolic}, in which LLMs serve as translators that convert user text input into symbolic representations, which are then processed by symbolic engines with strong capabilities in explanation, verification, and formal reasoning. Realizing this potential requires addressing the challenges identified in this work, such as ambiguities in linguistic spatial descriptions, conceptual neighborhood relationships in topological reasoning, and the correct use of available GIS functions and analytical workflows.
Additionally, our dataset is currently limited to the city and state levels, and further investigation into multi-scale spatial relations is still needed to fully capture the complexity of spatial interactions across different geographical scales and heterogeneous datasets. 
Moreover, the scope of this work is limited to topological relations in natural languages. While we can handcraft datasets with ground truth labels of formalized predicates for evaluation, the mathematical computation makes LLMs less competent than spatial databases. Directional and distance relations introduce more vagueness in language use due to factors such as shape and scale, especially when two places cannot be viewed as points. In future work, we plan to explore other spatial relations using datasets such as the Geograph project \footnote{http://www.geograph.org.uk}, which provides rich expressions of various spatial relations associated with geometries, text descriptions and photos ~\citep{m2011interpreting}. This will enable us to evaluate LLMs' or multi-modal foundation models' capabilities (e.g., vision-language geo-foundation models) in geospatial reasoning from a more comprehensive perspective~\citep{mai2023opportunities}. 
Lastly, we rephrased our own text instead of directly using paragraphs from DBpedia to allow for flexibility in introducing different context information. However, this approach may result in some loss of authenticity in language use, such as anaphora, which is prevalent in original text documents and worth exploring in future research.

In summary, this research demonstrates the promise and limitations of using state-of-the-art LLMs to analyze topological spatial relations, while offering insights for future research of advancing LLMs with geographical knowledge, aiming to develop GeoAI foundation models capable of qualitative spatial reasoning and other spatial intelligence tasks.

\section*{Disclosure statement}
The authors report there are no competing interests to declare.

\section*{Data and Codes Availability Statement}
The data and codes supporting the main findings of this study are available at Figshare: \url{https://doi.org/10.6084/m9.figshare.25127135.v1} and the GitHub repository at \url{https://github.com/GeoDS/GeoFM-TopologicalRelations}.

\section*{Acknowledgments}
Song Gao acknowledges the funding support from the National Science Foundation funded AI institute [Grant No. 2112606] for Intelligent Cyberinfrastructure with Computational Learning in the Environment (ICICLE). Any opinions, findings, and conclusions or recommendations expressed in this material are those of the author(s) and do not necessarily reflect the views of the funder(s).

\section*{Notes on contributors}

\noindent \textbf{Yuhan Ji}: Yuhan Ji is a PhD student GIScience at the Department of Geography, University of Wisconsin-Madison. Her main research interests include transportation, geospatial data science, and GeoAI approaches to human mobility. 

\noindent \textbf{Song Gao}: Dr. Song Gao is an associate professor in GIScience at the Department of Geography, University of Wisconsin-Madison. He holds a Ph.D. in Geography at the University of California, Santa Barbara. His main research interests include GeoAI, geospatial data science, spatial networks, human mobility and social sensing.

\noindent \textbf{Ying Nie}: Ying Nie is an undergraduate student at the Department of Computer Sciences, University of Wisconsin-Madison. Her main research interests include geospatial data science and GeoAI. 

\noindent \textbf{Krzysztof Janowicz}: Dr. Krzysztof Janowicz is a Professor in Geoinformatics at the Department of Geography and Regional Research, University of Vienna. His research interests are knowledge representation and reasoning as they apply to spatial and geographic data, e.g. in the form of knowledge graphs.

\noindent \textbf{Ivan Majic}: Dr. Ivan Majic is a Postdoc researcher in Geoinformatics at the Department of Geography and Regional Research, University of Vienna. His research interests include Spatial Data Science, GeoAI, and Qualitative Spatial Reasoning. 

\noindent

\bibliographystyle{apalike}
\bibliography{references}

\end{document}